\documentclass{styles/svproc}

\usepackage{url}

\usepackage[numbers]{natbib}
\usepackage{graphicx}
\usepackage{textcomp}
\usepackage{xcolor}
\usepackage{float}
\usepackage{caption}
\usepackage{subcaption}
\usepackage{amsmath}
\usepackage{amssymb}
\usepackage{authblk}
\usepackage[english]{babel}

\usepackage{soul,color}
\usepackage{enumitem}  

\usepackage{rotating, graphicx}
\usepackage{csquotes}

\begin{document}
\mainmatter              

\title{Self-awareness in intelligent vehicles: Feature based dynamic Bayesian models for abnormality detection}

\titlerunning{Self-awareness in IV: Feature based abnormality detection}

\author{Divya Thekke Kanapram \textsuperscript{1,3}, Pablo Marin-Plaza \textsuperscript{2},  Lucio Marcenaro \textsuperscript{1}, David Martin, \textsuperscript{2} Arturo de la Escalera \textsuperscript{2}, Carlo Regazzoni\textsuperscript{1}}

\institute{\textsuperscript{1}Department of Electrical, Electronics and Telecommunication Engineering and Naval Architecture, University of Genova, Italy. divya.thekkekanapram@edu.unige.it, \{carlo.regazzoni,lucio.marcenaro\}@unige.it\\ %
\textsuperscript{2}Intelligent Systems Lab, Universidad Carlos III, Leganes Spain. \{pamarinp, dmgomez, escalera\}@ing.uc3m.es\\ %
\textsuperscript{3}Centre for Intelligent Sensing, School of Electronic Engineering and Computer Science (EECS), Queen Mary University of London, UK.\\%
}


\authorrunning{D.T. Kanapram, P. Marin-Plaza, L. Marcenaro et al.}

\maketitle              
\begin{abstract}

The evolution of Intelligent Transportation Systems in recent times necessitates the development of self-awareness in agents. Before the intensive use of Machine Learning, the detection of abnormalities was manually programmed by checking every variable and creating huge nested conditions that are very difficult to track. This paper aims to introduce a novel method to develop self-awareness in autonomous vehicles that mainly focuses on detecting abnormal situations around the considered agents. Multi-sensory time-series data from the vehicles are used to develop the data-driven Dynamic Bayesian Network (DBN) models used for future state prediction and the detection of dynamic abnormalities. Moreover, an initial level collective awareness model that can perform joint anomaly detection in co-operative tasks is proposed.

The GNG algorithm learns the DBN models' discrete node variables; probabilistic transition links connect the node variables. A Markov Jump Particle Filter (MJPF) is applied to predict future states and detect when the vehicle is potentially misbehaving using learned DBNs as filter parameters.

In this paper, datasets from real experiments of autonomous vehicles performing various tasks used to learn and test a set of switching DBN models.


\end{abstract}

\section{Introduction}\label{section I}
Due to the rise of the population in the cities, the vehicles grow exponentially, leading to congestion and pollution. Consequently, road accidents increase dramatically due to numerous reasons such as lack of contextual data, distracted and reckless driving, adverse weather conditions, animal crossing, unsafe lane changes, etc. These factors show the importance of making the vehicles "self-aware" in order to ensure safety in driving. The future autonomous vehicles will use these self-awareness processes to solve different tasks in different environments that never were considered previously for decision making or problem-solving, such as trajectory planning or collision avoidance, or abnormality detection. The new learning capabilities from observing the environment or the autonomous vehicles' own states of behavior create new possibilities for problem-solving in new situations of the real traffic scenarios. So, developing self-aware models will improve the general decision and the navigation in the autonomous vehicles facilitating the improvement of the incremental self-capabilities, such as fault-tolerant decisions based on own perception or communication capabilities in dynamic environments.

Self-awareness (SA) is a broad concept that describes the cognitive property of an agent. In the case of artificial agents like intelligent vehicles (IVs) \cite{xiong2010autonomous}, the concept of SA is an ability to observe themselves and the surrounding environment through the various exteroceptive and proprioceptive sensors and process the sensory data to learn and maintain a contextual representation of the system \cite{regazzoni2020multisensorial}. Nowadays, the emergent techniques and algorithms in Machine learning allow for the learning of data-driven models that can provide self-awareness functionalities. Self-awareness functionality can be extended to 'collective self-awareness' by utilizing the shared data between the agents and learn models from the multi-sensory data collected from agents performing co-operative tasks. However, the in-depth analysis of communication schemes is beyond the scope of this paper. Here we have shown the co-operative communication scheme used in the experimental scenarios to collect data sets and an initial level collective self-awareness model to represent agents' joint tasks. 

Dynamic Bayesian Networks(DBNs) are probabilistic models representing the multi-sensory temporal data sequences at different abstraction levels \cite{murphy2002dynamic}. In this work, we have used a specific category of DBN models called Switching Linear Dynamic Systems (SLDS) \cite{fox2009nonparametric}. In SLDS, a sequential combination of linear dynamic models can be used to represent a non-linear dynamic model. The discrete variables in the higher levels of the SLDS represent switching variables (or superstates). These switching variables are associated with the linear dynamic models defined in the DBN's continuous state level.

The proposed methodology for learning the data-driven models for agents' self-awareness functionality has been verified with the data sets collected from autonomous vehicles performing co-operative and individual driving tasks of different scenarios.   

The main contributions of this paper can be summarized as follows:\\
1. A method to learn switching Dynamic Bayesian Network (DBN) models from the pairs of exteroceptive and proprioceptive sensory data sequences. Such learned models can detect abnormal behaviors in real-time (online phase) by testing the models with the data sets extracted from different experiences.

2. Performance analysis of the pair based DBN models was conducted to check the best pair based feature to detect abnormal situations in the surrounding environment. Considered the data sets collected from co-operative driving scenarios along with a scenario consists of one agent. 

The remainder of this paper is organized as follows. Section 2 presents a survey of the related work. In section 3, the proposed method is described by defining principles exploited in the training phase and the anomaly detection steps involved in the model the test phase. Section 4 summarizes the experimental setup in addition to the description of the research platform used. Section 5 presented abnormality measurement results of model testing, made analysis, and comparison. Finally, section 6 concludes the paper by including possible future research lines.

\vspace*{-\baselineskip}
\section{State of the art} 

This section explains some of the related work regarding the development of self-awareness in \textit{agents}. 
Over many years, self-awareness has been studied in multiple research disciplines, such as cognitive sciences, psychology, and philosophy \cite{bajgar2005development,asendorpf1996self, baker1897identification,tawney1902feeling}. The concept of self-awareness is widely studied in biology, which has been reproduced in artificial systems to enrich the capability of autonomy in different fields, including machine learning and robotics \cite{rinner2015self,Winfield2014237}. Moreover, in \cite{duval1972theory, goukens2009me}, the different aspects of self-awareness are discussed. \\
Here we consider DBNs learned using sensorial data recorded during training experiences as generative models capable of allowing a SA agent to predict states in future similar testing experiences. Additional probabilistic inference features related to DBN models allow the SA of the agent to detect possible abnormalities in new experiences. Prediction, estimation, and abnormality detection are the emergent (i.e., data-driven) SA features discussed in this paper as a collective property of a set of agents aiming to perform the same task. 

Developing self-awareness in agents has been shown to help an agent increase confidence when executing autonomously tasks and explainability of its own actions in terms of emergent SA models learned. Human efforts in different areas can take advantage, and in some fields, this can increase confidence in autonomous systems so allowing the human to reduce its overcontrol work.  An Intelligent vehicle \cite{bishop2000intelligent} can be seen as a straightforward example of an agent: it perceives information from the surrounding environment and uses obtained information to make decisions autonomously in different situations. However, this does not always imply that the IV can explain to itself and to the human user the causal sequence of events that carried it to make decisions. Self-awareness addresses within Artificial Intelligence the set of techniques/models that allow agents/machines to describe the relationship between perceptions and actions the agent has to do to perform a task. In this context, Machine Learning and Deep Learning, etc. are increasingly used to obtain SA models in a data-driven way, and self- driving vehicles \cite{berger2014competition} can benefit from such methods. Machine Learning techniques capable of dealing with uncertainty to learn the SA model from multisensory signals coming from the vehicle’s sensors are particularly useful. Such models can be of the generative type, so allowing predictions of future or lower-level states of the agent in consideration to be made when analyzing new data sequences.

In recent years, the research in intelligent and autonomous vehicles occupies a prominent place in the field of ITS. 
In \cite{8462490}, the authors propose an approach to develop a multilevel self-awareness model learned from an agent's multisensory data. Such a learned model allows the agent to detect abnormal situations present in the surrounding environment. In another work \cite{8455667}, the learning of self-awareness models for autonomous vehicles is based on the data collected by different maneuvering tasks performed by a human driver. In this work, visual perception and position data are used as modalities, and the cross-correlation between different modalities is analyzed for detecting abnormal situations. On the other hand, in \cite{article}, the authors propose a new architecture for mobile robots with a model for risk assessment and decision making when detecting difficulties in the autonomous robot design. In \cite{xie2018driving}, the authors proposed a model of driving behavior awareness (DBA) that can infer driving behaviors such as lane change. 
In \cite{8767204}, an approach to detect abnormalities in dynamic systems by the models learned from the different features of an \textit{agent} is presented. Moreover, it examines the most precise model to detect abnormal situations. However, it involves one agent and doesn't highlight the issue of co-operative driving. \\
In most of the related works, either the data from one entity is used, or the objective was limited; for example, in \cite{xie2018driving}, the aim was to detect lane changes either on the left or right side of the considered vehicles. In this work, we have considered the data from the real vehicles and developed pair based switching DBN models for each vehicle and, finally, obtained the performance comparison.
Moreover, shown an initial level collective DBN model and performance comparison has been made with the results obtained from single pair based DBN models. 
\vspace*{-\baselineskip}
\section{Proposed method}\label{section II}
This section discusses how to model \enquote{intelligence} and \enquote{awareness} into vehicles to generate \enquote{ Self-aware, intelligent vehicles.} Such intelligent vehicles should be provided of sensors to perceive internal as well as external states. Intelligence in this context is often related to the capability of using perceptions to allow the vehicle to adapt to variations in external situations. However, the system cannot be considered self-aware until the vehicle is not provided with the capability to observe perception and actions it has successfully done until a given moment. Then, organize them in a data-driven way into models capable to statistically predict at different hierarchical levels future states in case a similar situation and task to perform once again. This prediction capability should be inherent to the SA model, which has to be capable of deriving conditional models based on available knowledge (e.g., state at a previous instant) to derive predictions (e.g., future state). Models that have this capability are called generative models \cite{jebara2012machine}. 

The concept of anomaly arises when the SA model has a further inference capability: it can estimate whether current observations are in line or not with predictions of variables at whatever level in the model. This work uses this framework to define abnormality:  a dynamic anomaly is found by a SA model when the states the model predicts mismatch the current sensory observations. In this sense, the emergent self-awareness property in vehicles derives from two aspects: 1) to learn generative SA models from normal sensor experiences (Dynamic Bayesian Network (DBN) models are here used); 2) to associate to such models a general inference mechanism capable of performing predictions and of detecting dynamic anomalies. The latter capability can be defined as self-evaluation of models available in the vehicle: if a given SA model has the capability of predicting the future states efficiently, i.e., the ground truth observations from the vehicle’s sensors confirm predictions, than the vehicle is ensured that a certain level of homeostasis (dynamic stability) will occur in next moments performing the same actions it did in past experiences.

The capability of understanding when homeostasis conditions cannot be met with available models, i.e., anomaly detection, is so an important self-awareness step to allow an agent to start making decisions based on different strategies than normal known ones. 
The intelligent vehicles used in this work are equipped with one lidar of 16 layers and 360 degrees of Field of view(FOV), a stereo camera, and position/control encoder devices to make available information about the internal and position state during different tasks being performed. In this work, we mainly concentrated on such latter low dimensional multi-sensory observed data: namely the position (odometry) and control signals of the vehicle (e.g., steering, power, and rotor velocity ). In this work, position data are considered exteroceptive information as the relative position is used with respect to the environment around it. This information can be obtained, e.g., by GPS or by Localization techniques using exteroceptive sensors like lidar and cameras. However, here we consider as already available in low dimensions (coordinates with respect to environment) such data as extracted by sensor ad hoc algorithms.  On the other hand, the agents' control signal is here used as an example of proprioceptive sensory data.

Considering directly low dimensional data sensor observations makes it easier here to analyze how to learn pair-based switching DBN models, i.e., the generative models here used, and to associate anomaly detection tools to them. Extension to higher dimensional data (i.e., directly mapping lidar and video cameras observations as random variables into DBNs) requires methods based on probabilistic versions of tools like Variational Autoencoders(VAEs) and Generative Adversarial Networks(GANs) to be studied in detail to allow joint learning of mapping of high dimensional sensory observations into DBN states and prediction models as shown in \cite{9122766,8451418}. The extension to such high dimensional sensors of the SA model goes beyond the scope of this paper. Instead, here, we use low dimensional sensory data to explore the simultaneous SA awareness onto multiple vehicles jointly cooperating in a task. This case shows how collective awareness can rise as a direct extension of appropriate self-awareness models of individual agents in each vehicle. 

The proposed method is divided into two phases: offline training and online testing. A block diagram representation is shown in Fig. 1. In the offline training phase, the vehicles learn probabilistic filtering models from the multisensory data sequence while they perform a reference situation task. Each vehicle learns a set of models from the different combinations of the vehicle's control and position information. In the online test phase, each DBN model inside a vehicle is used as parametric knowledge of a  filter that makes inference on the corresponding sensory data. Each filter is provided with an additional computational method that allows it to estimate its own level of fitness(anomaly detection)  by comparing the states it can generatively predict with the sensors' current observations.  The abnormality occurs in the environment are detected as deviations from expected modeled behaviors. A filter called Markov Jump Particle Filter (MJPF) \cite{costa2006discrete,kanapram2019self} is applied to the learned DBN models to perform the filtering operation of prediction and update. The filter is modified to allow it to also estimate abnormalities at different abstraction levels as a required  SA capability.
\vspace*{-\baselineskip}

\subsection{Offline training phase} \label{offlinetr}

In this phase, the considered agents (i.e., vehicles) learn the vocabulary and dynamic/hierarchical models of the switching DBN  from the observed exteroceptive and proprioceptive sensorial data that originate from the current agent’s experience defining their normal behavior.

DBN models are of a generative type, and learning them implies learning the conditional probabilities that relate their variables.   Therefore, observing a sequence of data and learning conditional functions within the DBN will allow the agent to predict variables not yet observed based on evidence collected until that moment. This makes the model appropriate for capturing causal relationships at the basis of SA.
 
\begin{sidewaysfigure}
\includegraphics[width=19cm,height=20cm,keepaspectratio]{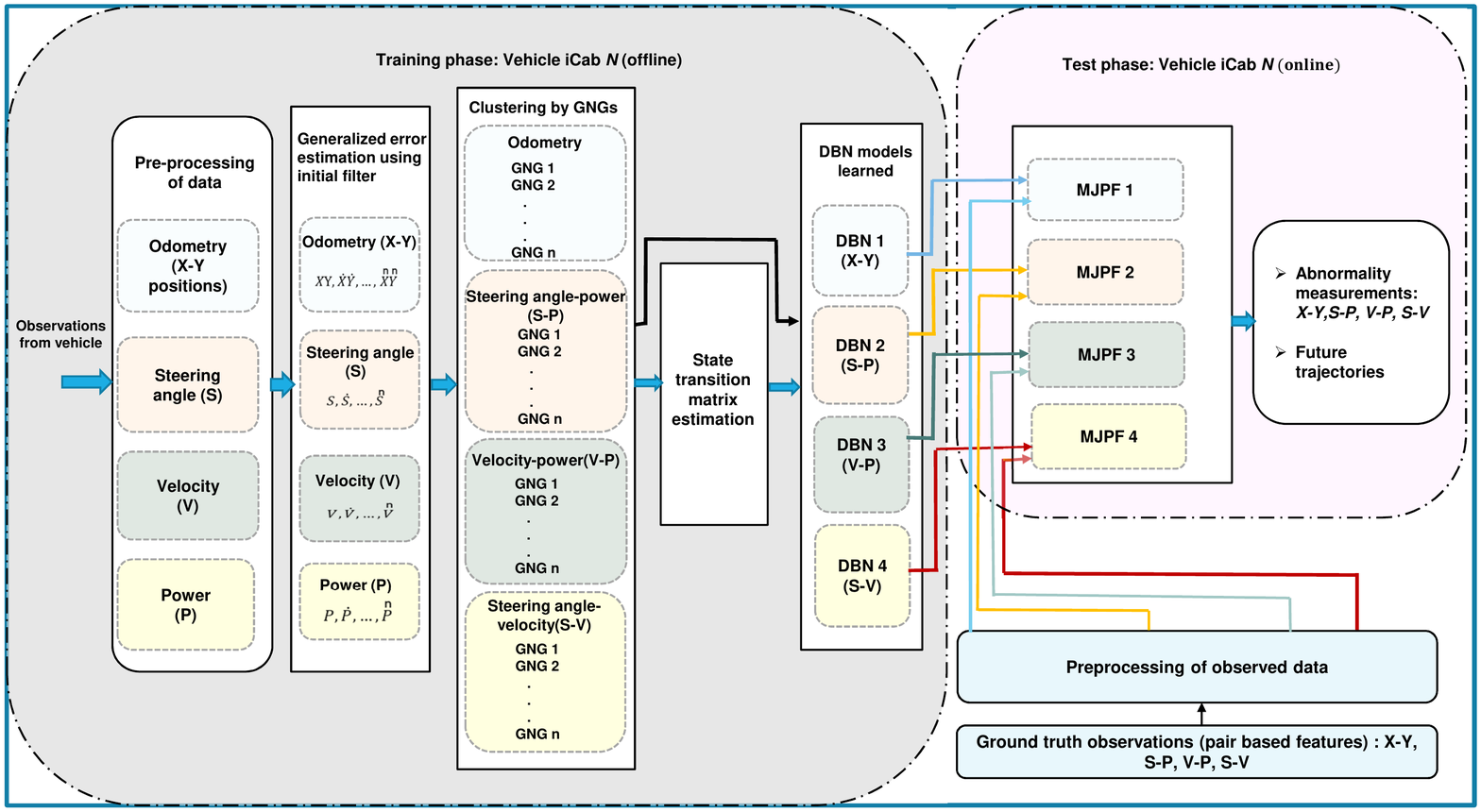}
\caption{Block diagram: training and test phase of switching DBN model}
\label{fig:blockdiagram}
\end{sidewaysfigure}

The grey shaded area in Fig \ref{fig:blockdiagram} shows the block diagram representation of the training phase. The various steps involved in switching DBN model learning have been described below.

\subsubsection{Estimation of Generalized states}~\\
The method assumes a synchronization operation is available on the considered multisensory data. Exteroceptive based position data and three pairs of vehicle control signals are considered in this paper as a case study. An initial generalized filter \mbox{\cite{campo2017static}} is applied to the sequence of data variables to estimate so-called generalized states (GSs). Generalized states are the variables here used to describe anomalies with respect to to a given DBN model. Using GSs allows the agent to estimate a new DBN model capable of predicting in a better way in future those anomalies coming from sequences statistically similar to the one that has generated such anomalies.  Generalized states are generalized coordinates that allow each variable in the DBN to describe its own dynamics as part of the random information. Therefore, conditional probabilities in a DBN  formed by GS variables allow predictions of the dynamic evolution of states such variables represent with advantages in the inference process.      

The  generalized state of a continuous state variable associated with a given sensory data $c$ can be defined as:  

\begin{equation}\label{eq1}
\boldsymbol{\tilde{X}_{t_k}^c} = [X_{t_k}^c \hspace{2mm}  \dot{X}_{t_k}^c \hspace{2mm} \ddot{X}_{t_k}^c  \hspace{0.5mm} \cdots \hspace{0.5mm} X^{c,(L)}_{t_k}]^\intercal,
\end{equation}
where $(L)$ indicates the $L$-th time derivative of the state. \\
The $l$-th time derivative of GS at the time $t_k$ can be written as: 

\begin{equation}\label{eq2}
X^{(l)}_{t_k} = \frac{X^{(l-1)c}_{t_k} - X^{(l-1)c}_{{t_k}-1}}{\Delta {t_k}},
\end{equation}

where $X^{(0)}_{t_k} = X_{t_k}$ and $\Delta {t_k}$ is the uniform sampling time for all variables considered.
In this work, the derivatives inside the generalized state(GS) are limited L=1, i.e., only the state and its derivative are included in the GS. This is because we used a DBN with limited memory (a two-slice DBN) so that dynamical models depend directly only on variables in the previous slice. In the case of low dimensional observations (and consequently states), this implies that local interslice dynamic models can be considered as well approximated by linear equations between GSs. So the model learned with GSs can capture the considered agent dynamics as piecewise linear. DBNs with longer memories imply higher-order derivatives so that additional learning tools would be required with respect to those here discussed. In the future, the work here presented could be extended by including higher-order GSs.

In a vehicle agent, an initial DBN has to be assumed as known that embeds basic dynamic conditions. In this case, a DBN is used that assumes that there will not be any force acting on the agent between two consecutive time instants but random Gaussian perturbations. The corresponding filter can be defined in terms of GSs:  the state vector at $t_{k+1}^{th}$ time instance will remain same w.r.t the state vector at $t_{k}^{th}$. So that the filter produces derivatives of states as errors. So GSs estimated by such an initial filter represent jointly for each given state X,  errors, i.e., derivatives of X.  GSs can therefore be seen as pairs (state, error) that are produced in this case by an initial filter associated with initial DBN.
Mainly we have considered four pairs of GSs data for DBN model learning, such as odometry($X-Y$), steering-power ($S-P$), velocity-power ($V-P$), and steering-velocity ($S-V$).
\subsubsection{Construction of discrete level of DBN model}~\\
This section explains the process involved in learning the vocabularies of discrete level variables, as well as dynamic models, transition probabilities, and co-occurrence matrices between variables of the DBN model, i.e. all conditional probabilities that can decompose the joint probability of the DBN in a set of causally meaningful conditional probabilities, making it a generative model.

\begin{figure}
\centering
 	\includegraphics[width = 1 \linewidth ]{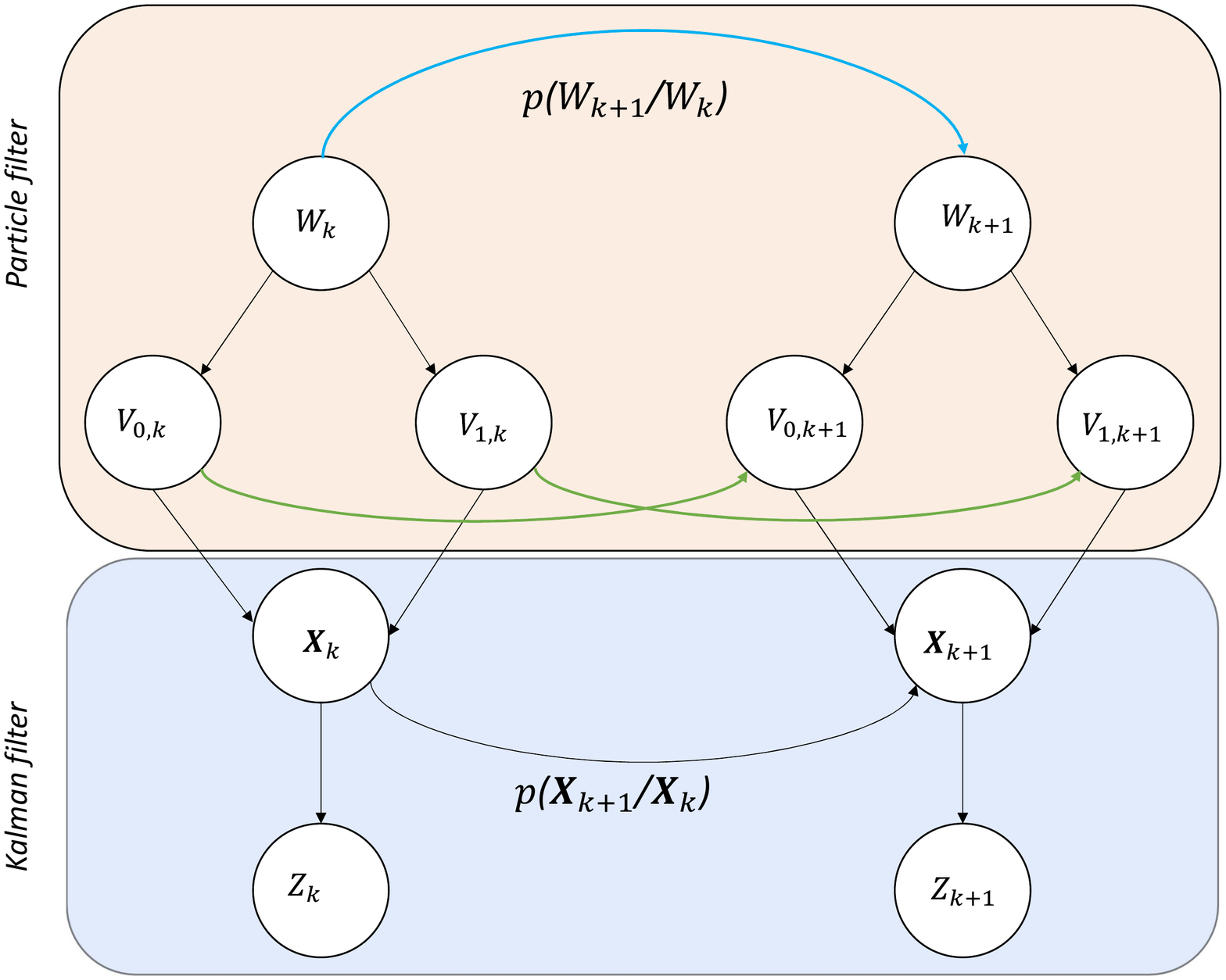}
\caption{Proposed switching DBN model}
\label{fig:DBN}
\end{figure}

This work considers a hierarchical switching 2-Time Slice DBN (2T-DBN)\cite{Koller2001SamplingIF} as a generative filter that can predict future states of the considered entities (i.e., vehicles). Fig.\ref{fig:DBN} shows the representation of the DBN model, continuous level constitutes the generalized states, and discrete level represents the hierarchically higher semantic vocabulary and correlation among them. Each of the discrete level variables belongs to a particular dynamical model in the continuous state space. Thus, the considered switching DBN model can represent non-linear dynamics by a set of linear dynamic models. 

To model the discrete level of the DBN (orange shaded area in Fig. \ref{fig:DBN}), the discrete switching variables that represent the meaningful information have to be learned from the generalized states estimated from the outcomes of the initial filter on a given data sequence. A clustering algorithm called Growing Neural Gas (GNG) \cite{fritzke1995growing} is used to group generalized states samples collected as output of a previous filter associated with a DBN  (in this case, the initial filter). Each cluster of samples defines a different switching variable of the newly learned DBN. 

In general, clustering can be described as the process of organizing a collection of k-dimensional vectors into groups whose members share similar features in some way. A k-dimensional vector represents each one of such groups called a code vector (other names used are centroid and node). 
There are many algorithms available for clustering: K-means \cite{macqueen1967some}, Self Organising Map(SOM) \cite{kohonen1990self}, Neural Gas(NG) \cite{martinetz1991neural}, Growing Neural Gas(GNG) \cite{fritzke1995growing}, Density-based spatial clustering of applications with noise (DBSCAN)\cite{ester1996density}, etc. The SOM algorithm can compress large multidimensional datasets into a fixed number of representative units. However, the dimension of the representative units (clusters) needs to be defined before, and it may sometimes cause not intuitive for representing the characteristics of data structure.

In contrast to the SOM, GNG is an unsupervised, adaptive, and incremental neural network that learns topologies; it grows during the learning process and does not require users to define the number of representative units called nodes beforehand.  Such a dynamic property is an advantage over other clustering algorithms for using it in many applications. DBSCAN is a density-based clustering algorithm that finds a number of clusters starting from the estimated density distribution of corresponding nodes.  Although it has many advantages, such as discovering arbitrarily shaped clusters and robust detection of outliers, the algorithm sometimes fails to identify clusters in those situations of varying density of considered data or if the dataset is too sparse. The dataset considered in this work is sparse and multidimensional so that we have chosen the GNG algorithm by considering its advantages over other clustering algorithms.

The GNG produces a set of nodes, and each of the nodes represents a cluster.  The cluster can be described in terms of global statistical properties of the samples grouped in it:  mean and co-variance is parameters here used. In the proposed method, GSs components (i.e., state and errors)  ( Eq.\ref{eq1}) obtained from the initial filter are separately clustered by the GNG. For  example, in the case of control $S-P$ modality clustering is applied separately to GSs components  written as below:
\begin{subequations}
  \begin{equation}
    \label{eq-3a}
    GNG1 = [s_{t_k} \hspace{2mm}  p_{t_k} ]^\intercal
  \end{equation}
  \begin{equation}
    \label{eq-3b}
  GNG2 = [\dot{s}_{t_k} \hspace{2mm} \dot{p}_{t_k}]^\intercal
  \end{equation}
\end{subequations}

In GNG, each node groups a subset of data samples that are closer with respect to a defined distance metric to the centroid of the node. The nodes obtained by each GNG define a set of \textit{letters} that can be used to describe the new DBN to be learned. The collection of nodes generated by the GNGs of modality $S-P$ can be written as below:

\begin{subequations}
  \begin{equation}
    \label{eq-a}
    {V_{SP}^{0}}= \{{a_{1},a_{2}, ..., a_{p}}\}
  \end{equation}
  \begin{equation}
    \label{eq-b}
   {V_{SP}^{1}} = \{b_{1},b_{2}, ..., b_{q} \}
  \end{equation}
\end{subequations}

where $p$ and $q$ represents the index of clusters obtained by the state and derivative GNGs, $V_{SP}^{0}$ and $V_{SP}^{1}$ represent the group of letters(nodes) produced by GNG1 and GNG2 respectively.

Each GNG produces clusters (i.e., vocabularies of letters) that have to be coupled to define a single dynamic behavior: to define how to couple letters describing states and derivatives; time co-occurrence is used.  Clusters of states and derivatives that are activated simultaneously are associated with a pair of letters to generate different switching variables for the DBN model.  
Each pair can form a \textit{word} and an example belong to $ S-P $ modality is provided below:

\begin{equation}\label{eq3}
{W_{i}^{SP}} = [V_{m}^{0},V_{n}^{1}]
\end{equation}

where $V_{m}^{0}$ represents the $m^{th}$ element of the set of nodes generated by GNG1 and $V_{n}^{1}$ is the $n^{th}$ element of node belong to GNG2. Each pair can be called as a \textit{word}, and a unique label is assigned to each \textit{word}. The list of words, along with assigned labels, form a dictionary as below.

\begin{equation}\label{eq4}
{D_{sp}} = \{\beta, {\beta_1}, \cdots  \beta_L\}
\end{equation}

where ${\beta_l}$ is ${W_l}$. $G_l$ and $G_l$ represents the unique label assigned for word $W_l$.
Dictionary is a high-level hierarchy switching variable that explains the system states from a semantic viewpoint. 

Furthermore, the transition probability matrices have been estimated based on the timely evolution of such \textit{letters}. The transition matrix consists of the probability of activating nodes from state and derivative space at the next time instance by knowing the currently activated nodes. The \textit{green} and \textit{blue} coloured arrows in Fig. \ref{fig:DBN} represents the transition probability between discrete variables. 

The model above can be extended to the case of collective awareness: when two GSs DBNs associated with two agents in a swarm in a collective ensemble are available. Collective awareness can be modeled as additional, conditional probabilities describing how one agent's GS can be predicted from another agent's GS.   As an example of how this additional, conditional probability can be represented by a  transition matrix between words of two different agents, the coupled co-occurrence matrix is here used that was obtained by considering the vocabulary (\textit{letters}) learned from the position data of agents. As in the previous case, firstly, each entity's position GSs have been separately clustered by GNG to generate respective vocabularies (i.e., \textit{letters}). Then, the co-occurrence matrix that encodes the probabilities of passing from \textit{words} describing    agent1 task to the \textit{words} describing  agent2 and vice versa.  Each element of the coupled co-occurrence matrix represents the probability of activation of one agent's discrete vocabulary(letters) conditioned to the other agent's GS clusters. The coupled co-occurrence matrix can be represented as below:

\begin{equation}
 \textrm{Coupled co-occurrence matrix}, T =
\begin{bmatrix} 
M_{11} & M_{12} & \dots    &M_{1n}\\
\vdots & \ddots & \\
 M_{m1} & M_{m2}  & \dots     & M_{mn}
\end{bmatrix}
\quad
    \label{eq:eqnmatrix}
\end{equation}

where $m$ and $n$ represent the maximum number of clusters (obtained from position data) belongs to agent1 and agent2, respectively. Each entry in the matrix represents the co-occurrence probability between agent1 and agent2. For example, $M_{12}$ (refer Eq. \ref{eq:eqnmatrix})the co-occurrence probability from node 1 (mean value of first cluster) belong to agent1 to node 2 (mean value of second cluster) belong to agent2. This information of couple probability helps to model the collective awareness for the agents jointly performing tasks. 

For each switching variable, a different dynamic model is learned that defines the dynamics of generalized states $\boldsymbol{\tilde{X}}_k$ in the region of the state covered by each switching word. As we limited GSs to L=1 each dynamic model can consider only clusters that collect as samples GSs including state and derivative, such that $\boldsymbol{\tilde{X}}_k = [X_k \hspace{2mm} \dot{X}_k]^\intercal$. For such a case, mean state (and related covariance) individuate sparse regions in the state space, while mean derivative (and its covariance) defines local linear dynamics around that state region. On this basis, for each switching variable in the word vocabulary, it is possible to learn  continuous dynamic models of random mean  velocity equal to the corresponding word for tracking generalized states' dynamics, such that:

\begin{equation}\label{eq5}
\boldsymbol{\tilde{X}}_{k+1} = A\boldsymbol{\tilde{X}}_{k} + BU_k +  w_k
\end{equation}
where
$$A = 
\begin{bmatrix}
    I_{j}   & 0_{j,j}   \\
    0_{j,j} & 0_{j,j}   
\end{bmatrix} \hspace{1mm} ; \hspace{1mm} B = 
\begin{bmatrix}
    0_{j,j}   \\
    I_{j}\Delta k    
\end{bmatrix}$$
The variable $j$ indexes the number of states in the data combination in consideration. $I_j$ is an identity matrix of dimension $j$. $0_{j,j}$ is a zero $j \times j$ matrix. $w_k \sim \mathcal{N}(0,\sigma)$, encodes the system noise. $U_k = E({W_{i,k+1}}|{W_{i,k}})$, where $E(\cdot)$ is the expectation operator. The control vector $U_k$ contains the agents velocity when it's state falls within a discrete space formed by the GNG clustering.

\subsubsection{Pair-based DBN models}~\\ 
All the previous steps are involved in learning the pair based switching DBN models from the multisensory position and control data. The number of DBNs learned by the vehicle $m$ can be written as:
\begin{equation}\label{eqdbn}
 \boldsymbol{DBN}m = \{DBN_1, \cdots, DBN_n\}.
\end{equation}

where $m$ represents the $m^{\text{th}}$ vehicle in the network and $n$ is the total number DBN learned by the $m^{\text{th}}$ vehicle. The same DBN architecture is considered for making inferences with different sensory data combinations belong to each vehicle considered. The learned DBN can be represented, as shown in Fig.\ref{fig:DBN}. The DBN has mainly two levels, such as continuous and discrete levels.

\subsubsection{Coupled DBN models}~\\
The coupled co-occurrence matrix learned from two agents' position data helps to model the interacting agents' joint/collective awareness. Each agent estimates the co-occurrence matrix in the training phase by jointly considering the agent1 and agent2 vocabulary(letters).
In the test phase, the intercommunication scheme proposed in Section \ref{section III} shares the observed position data between the agents at each time instance. With this information, the coupled DBN inside each agent estimates the discrete level coupled letters (i.e., \textit{word}). The coupled pair of letters at each time instance points to a specific cell in the co-occurrence matrix (refer Eq. \ref{eq:eqnmatrix}) to check the probability of firing the current \textit{word}. If the probability value is high, it means the agents' current experience matches with the one used to learn the model, and the situation is considered normal. When the value goes lower, the agents pass through unseen dynamics, and the situation is considered abnormal. The coupled DBN models are specially used in situations where the pair-based self-awareness DBN models fail to detect the anomaly happens around other agents in cooperative scenarios.
\subsection{Online test phase}
The block diagram representation for the online test phase is shown as a pink shaded area in Fig.\ref{fig:blockdiagram}. In this phase, a probabilistic switching model called Markov Jump Particle Filter (MJPF) \cite{1512727,8767204} has been chosen to make inferences on the DBN models (refer Fig.\ref{fig:DBN}) learned in the training phase. 

The filtering algorithms like  Markov Jump Particle Filter (MJPF) \cite{6000738,905890} and Interacting multiple models (IMM) filters \cite{4047965} allow an agent to predict and estimate target motion according to multiple probabilistic models. The filters differ in the inference methods they use to perform prediction and update steps. While MJPF uses particle filters at discrete levels together with Kalman filters at continuous levels, IMM filters can use different approaches. For example, in IMM filters  \cite{1299,8709675}, a  model-driven approach is performed to fuse Kalman Filters. In general, IMM filters can be coupled with parameter estimation learning methods specific to the inference approach used that can be used on training sequences. However, parameters are often chosen by design, and fixed discrete state transition probabilities are provided offline from the discrete variables switches' frequency. The number of models is generally a priori fixed, limiting the descriptors of the agents' dynamics. 
In this work, we used MJPF, a type of Markov Jump Linear System (MJLS) that uses a parametrized couple of Particle Filter and Kalman filters that can be learned from data. This allows as in IMM inferences on a Dynamic Bayesian Network jointly at continuous and discrete levels. However, the data-driven approach used in this work is based on a free energy minimization approach that allows a varying number of dynamic models to be estimated together with temporal transition probabilities that characterize the models' discrete temporal evolution. 

In IMM, model switching is mainly dependant on a time-independent transition probability matrix;  in MJPF here used, co-occurrence probability and transition models learned are time-dependent, so allowing a time-variant transition probability, specific for each dynamic model,  to be employed. In used MJPF, the number of particles employed at the discrete level is defined as proportional to the number of dynamic models. The method explore in parallel an alternative set of dynamic models predictions by evaluating the best choices depending on anomaly detection capabilities added.
\subsubsection{Estimation of future states and abnormality detection}~\\ 
The preprocessed data sequences collected from the experiences of agent vehicles not included in the training set are given as input to the MJPF in the online phase.
In a Markov Jump Particle Filter (MJPF), the posterior probability density function  can be written as:

\begin{equation}\label{eqmjpf}
 p({W_{k+1}},\boldsymbol{\tilde{X}}_{k+1}/ {Z_{k+1}}) = p(\boldsymbol{\tilde{X}}_{k+1}/{W_{k+1}},{Z_{k+1}})p({W_{k+1}}/{Z_{k+1}})
\end{equation}

where ${W_{k+1}}$ is the word in the higher hierarchical level and  $\boldsymbol{\tilde{X}}_{k+1}$ is the continuous state in the state space at time instant $k$. $ p({W_{k+1}}/{Z_{k+1}})$ is estimated by using  particle filter \cite{210672} at the discrete vocabulary level. Dynamic models learned in the training phase as in Eq. \ref{eq5} are used as a dynamic model equation of Kalman Filters at the continuous level associated with the discrete switching variable. A different gaussian velocity associated with the corresponding switching variable cluster obtained by GNG  is represented with the control vector as a variable $U_k$ (refer Eq. \ref{eq6}).  This means that the predictions $p(\boldsymbol{\tilde{X}}_{k+1}/\boldsymbol{\tilde{X}}_{k})$ of particles at the continuous state level are made by considering a bank of Kalman Filters built according to the discrete vocabulary where different Gaussian velocity models are valid for each switching variable. 

The goal is to use a filter capable of performing inference jointly at continuous and discrete levels(refer Fig. \ref{fig:DBN}). The MJPF allows the agent to predict its future states, starting from the learned DBN generative model's prediction components. Moreover, the filter is provided with an additional inference feature that allows it to use bayesian prediction and evidence messages exchanged by  DBN nodes to compute abnormality indicators at different DBN levels. Such abnormality measurements allow the agent to detect when its own prediction model components of the generative DBN fit with new data sequences processed by the modified MJPF. Anomaly is here defined as a time signal that indicates when the probabilistically predicted states mismatch the current noisy sensory observations. A detailed description of the modified  MJPF version with abnormality detection here used is described in section II of \cite{8455592}. However, in that paper, the authors used a different clustering algorithm(SOM) for estimating the switching variables associated with different dynamic models from generalized errors. Moreover,  a  different approach is used here in learning the discrete vocabulary and the correlation among them than the one used in \cite{8455592}.

In the prediction step of the MJPF, a Sequential importance Resampling (SIR) PF \cite{10.2307/2289460,gordon1993novel} is used to predict the new set of particles iteratively. The time-variant transition probability model is used to perform particle sampling at a new time instant at the discrete level.  Gaussian proposal function $ p({W_{k+1}}/{W_{k}})$ is used. In the update step, the new sample from the testing sequence is provided to the filter. It allows firstly to update the prediction message at the continuous level and to obtain the new posterior. The update is then propagated at the particle level to obtain new weights based on the observations that such posterior gives to the discrete variable itself. 
As each particle ${W_{k}}^{\ast}$,is associated with the  Kalman filter having the dynamic model of the particle  word $W_{k}$ than the corresponding dynamic model is used within the particle  to predict continuous state $p(\boldsymbol{\tilde{X}}_{k+1}/\boldsymbol{\tilde{X}}_{k},{W_{k}}^{\ast})$

The prediction performed at the continuous level can match with updates to a different extent. Measuring such a matching level by a probabilistic distance measurement allows the modified MJPF to estimate the anomaly that can be present. In probability theory, a statistical distance quantifies the distance between two statistical functions, which can be either two random variables or two probability distributions. Some important statistical distances used between two distributions include: Bhattacharya distance \cite{bhattacharyya1943measure}, Hellinger Distance \cite{pardo2005statistical}, Jensen–Shannon divergence \cite{endres2003new}, Kullback–Leibler (KL) divergence \cite{hershey2007approximating} etc.  Hellinger distance (HD) is a symmetric distance used to quantify the distance vectors having only positive or zero elements \cite{abdi2007encyclopedia}. Here such a distance is used to measure the distance between prediction and evidence at the continuous level.  The works in \cite{Lourenzutti2014} and \cite{kanapram2019self} also proposed to use Hellinger distance (HD) as an abnormality measurement.\\
 Accordingly, let $p(\boldsymbol{\tilde{X}}_{k}^c|\boldsymbol{\tilde{X}}_{k-1}^c)$ be the predicted generalized states and $p(Z_k|\boldsymbol{\tilde{X}}_{k}^c)$ be the observation evidence. The Hellinger distance (HD) can be written as:
\begin{equation}\label{eq6}
\theta_k^c = \sqrt{1 - \lambda_k^c},
\end{equation}
where $\lambda_k^c$ is defined as the Bhattacharyya coefficient \cite{Bhattacharyya1943}, such that:
\begin{equation}\label{eq7}
\lambda^c_k = \int \sqrt{p(\boldsymbol{\tilde{X}}_{k}^c|\boldsymbol{\tilde{X}}_{k-1}^c) p(Z_k^c|\boldsymbol{\tilde{X}}_{k}^c)} \hspace{1mm} \mathrm{d}\boldsymbol{\tilde{X}}_{k}^c.
\end{equation}
When MJPF processes a testing set sequences, an abnormality measurement is computed at each time instant at the continuous level, as in the equation \eqref{eq6}. The variable $\theta_k^c \in [0,1]$, where values close to $0$ indicate that the ground truth measurements match with predictions, whereas values close to $1$ reveal the presence of an abnormality in the environment. Once estimated the anomaly by HD metric, it is possible to check the complementarity among different DBN models learned by measuring how well the pair based models differ in performance of detecting environmental abnormalities.

In addition to the HD abnormality measurements, the co-occurrence of events with a low probability of occurring in coupled DBNs of the two agents provides a further collective abnormality measurement. This measurement can be used to measure the reciprocal influence of agent states in case of collective awareness. In this way, when an anomaly happens around one agent, other agents can understand this as a low probability co-occurrent event can happen. In this paper, we have explored how this works using only odometry data in this part of discrete level joint anomaly detection by coupled DBN model to provide an insight of possible future developments of the presented method. 
The coupled DBN model inside each agent first checks the firing of a pair of letters (word) formed from both agents' currently observed position data. Then record the probability value inside the corresponding cell of the coupled co-occurrence matrix. 
A high probability value means that the current experience matched the agent's experience in the training phase when the model was learned. A  low probability value detects the presence of a collective anomaly that is occurring to the agents' team. For the better representation, we have taken 1-(the value of co-occurrence probability) and plotted the resultant signal in Sec \ref{section IV}. This discrete level anomaly based on coupled co-occurrence matrix can be estimated with the following equation:
\begin{equation}\label{eqabn}
 \delta_{k} = 1-{M_{ij}^{k}}   
\end{equation}
where $M_{ij}^{k}$ is the probability value in the $i^{th}$ row and $j^{th}$ column of the co-occurrence matrix at $k^{th}$ time instance.

 This discrete level anomaly metric encodes the information of collective awareness. However, this part will require further work to improve the collective awareness functionality by considering the system's control data and other relevant features. In this paper, we have shown a preliminary level collective awareness model and the obtained results.
 
\section{Experimental Setup and employed datasets}\label{section III}
\begin{figure}
	\begin{subfigure}[t]{0.5\textwidth}
		\centering
		\includegraphics[width=5cm,height=4cm]{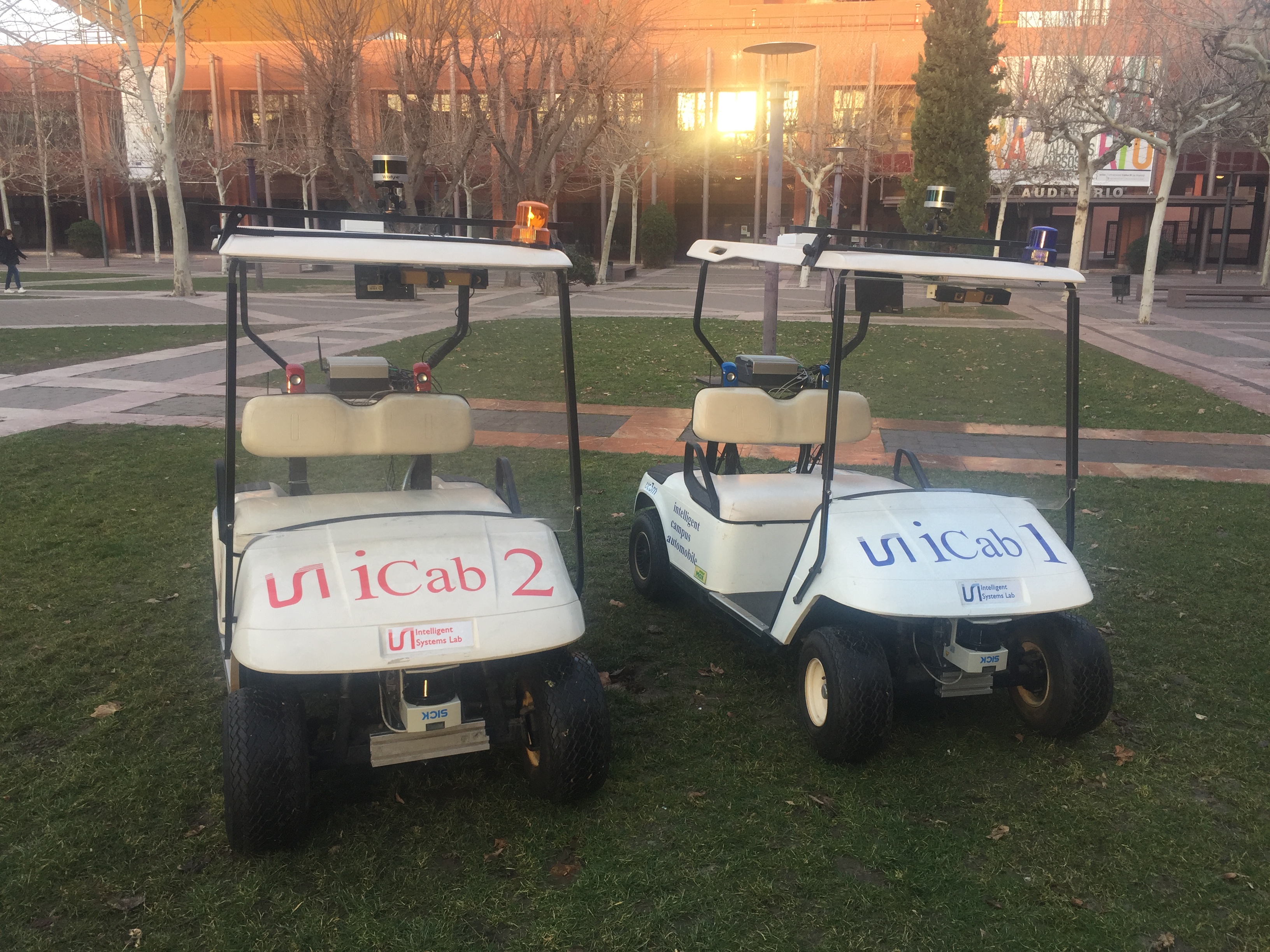}
		\caption{The autonomous vehicles (iCab)}
		\label{fig:iCab}
	\end{subfigure}%
	\begin{subfigure}[t]{0.5\textwidth}
		\centering
		\includegraphics[width=5cm,height=4cm]{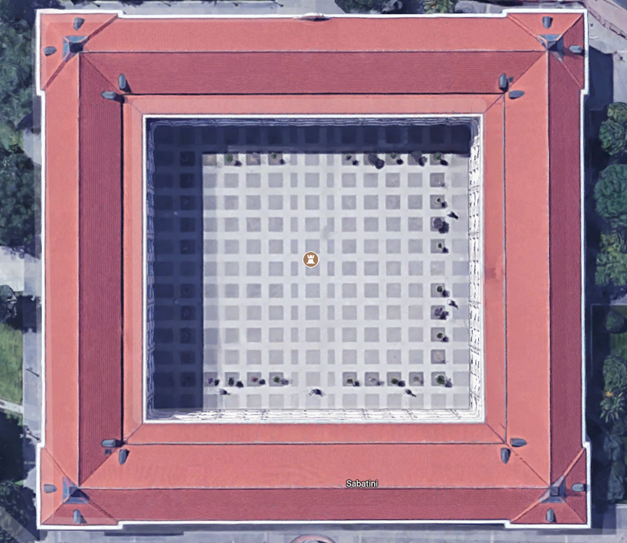}
		\caption{The environment}
		\label{fig:environment}
	\end{subfigure}
	\caption{The agents and the environment used for the experiments.}
	\label{fig:icabAndEnv}
\end{figure}

In order to validate the proposed method, the datasets collected from two intelligent research platform with autonomous capabilities called iCab (\textbf{I}ntelligent \textbf{C}ampus \textbf{A}utomo\textbf{B}ile)\cite{marin2018global} (see Fig.\ref{fig:iCab}) performing co-operative as well as individual driving tasks were considered. The multisensory data extracted from co-operative driving tasks used to learn the DBN models and test the models self/collective awareness functionality, whereas a single-vehicle scenario datasets exploited to learn and test DBN model's self-awareness capability. 

The considered iCab vehicles are equipped with two powerful computers and a screen for debugging and Human-Machine Interaction (HMI) purposes of navigating through the environment, as displayed in Fig. \ref{fig:environment}. The software prototyping tool used is ROS \cite{quigley2009ros}.
The exteroceptive position and proprioceptive control data sets collected from the \textit{iCab} vehicles performing tasks are firstly synchronized to match their time stamps. Additionally, the multisensory data normalized to bring the numeric columns in the data set to a common scale by not distorting the differences in the ranges of values or losing information.\\  
The intercommunication scheme proposed in \cite{kokuti2017v2x} was used to share all the position data and their respective timestamp information over the Virtual Private Network (VPN) during the co-operative driving experiments of \textit{iCab} vehicles. 

Both vehicles perform a Perimeter Monitoring Task (PMT) jointly , which consists of the autonomous movement of platooning around a square building (see Fig.\ref{fig:environment}). The 2D data of  exteroceptive odometry ($X-Y$) and the different pairs of the proprioceptive control variables such as Steering angle-Velocity ($S-V$), Steering angle-Power ($S-P$), and Velocity-Power ($V-P$) are the main features considered to learn and test the models. The dimension of the movement trace (Fig.\ref{fig:environment}) in the testing environment is 38mX33m.

\begin{figure}
	\begin{subfigure}[t]{0.48\textwidth}
		\centering
		\includegraphics[height=5cm, width=6cm]{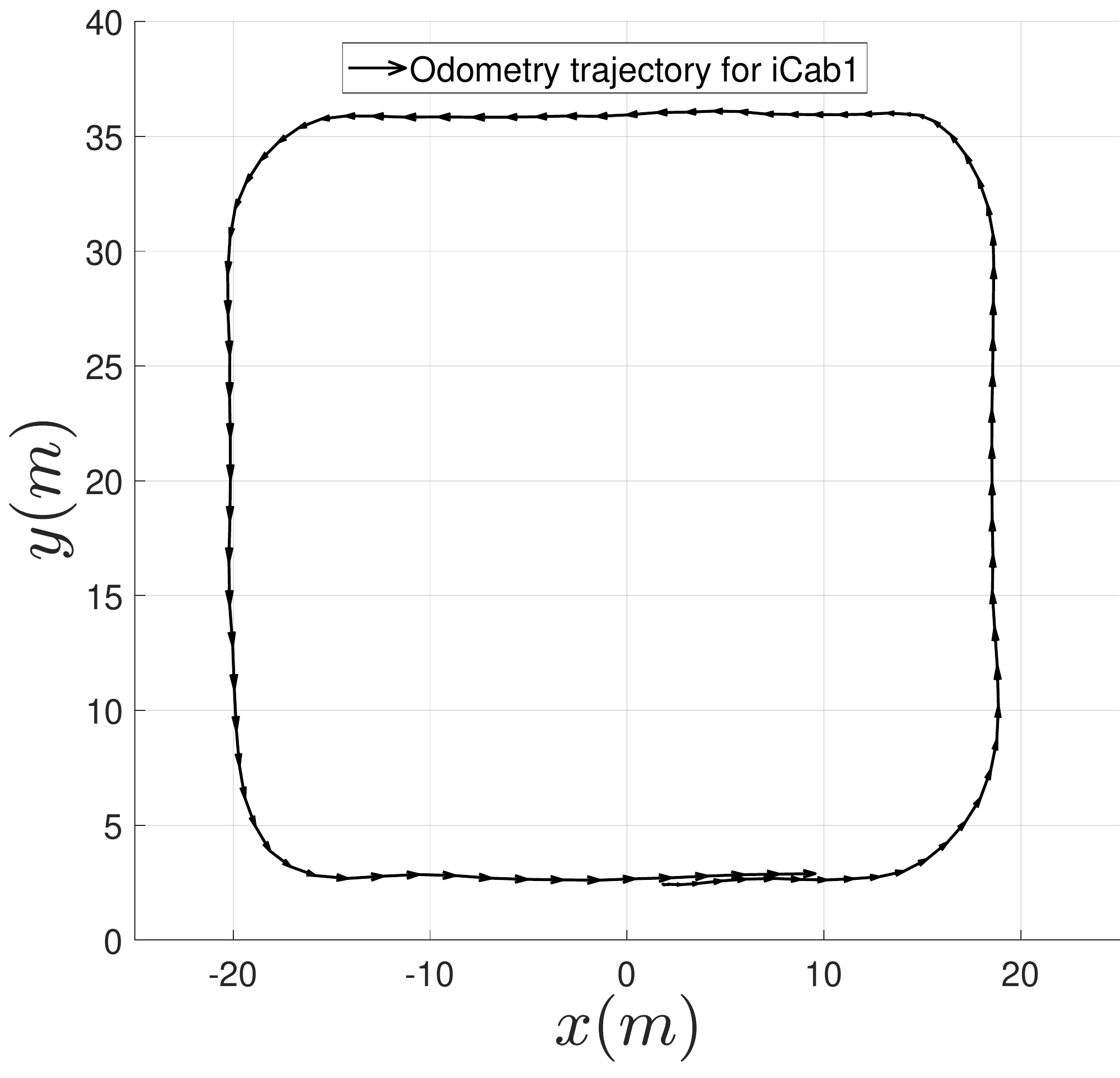}
		\caption{Perimeter monitoring}
		\label{fig:PM1}
	\end{subfigure}%
	\begin{subfigure}[t]{0.5\textwidth}
		\centering
		\includegraphics[height=5cm, width=6cm]{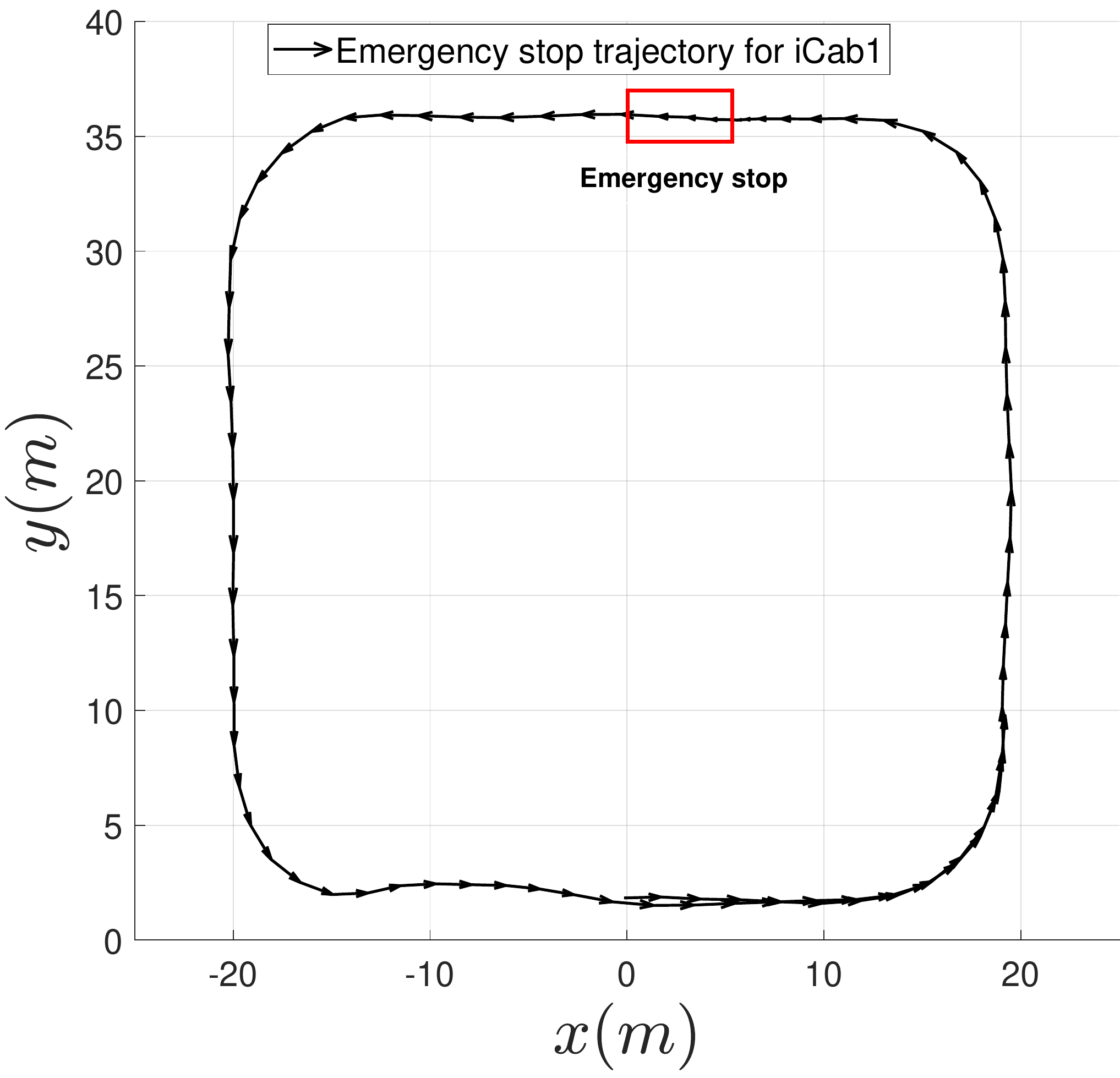}
		\caption{Emergency stop}
		\label{fig:ES1}
	\end{subfigure}
	\caption{Odometry data for iCab1}
	\label{fig:Position data iCab1}
\end{figure}

\begin{figure}
	\begin{subfigure}[t]{0.48\textwidth}
		\centering
		\includegraphics[height=5cm, width=6cm]{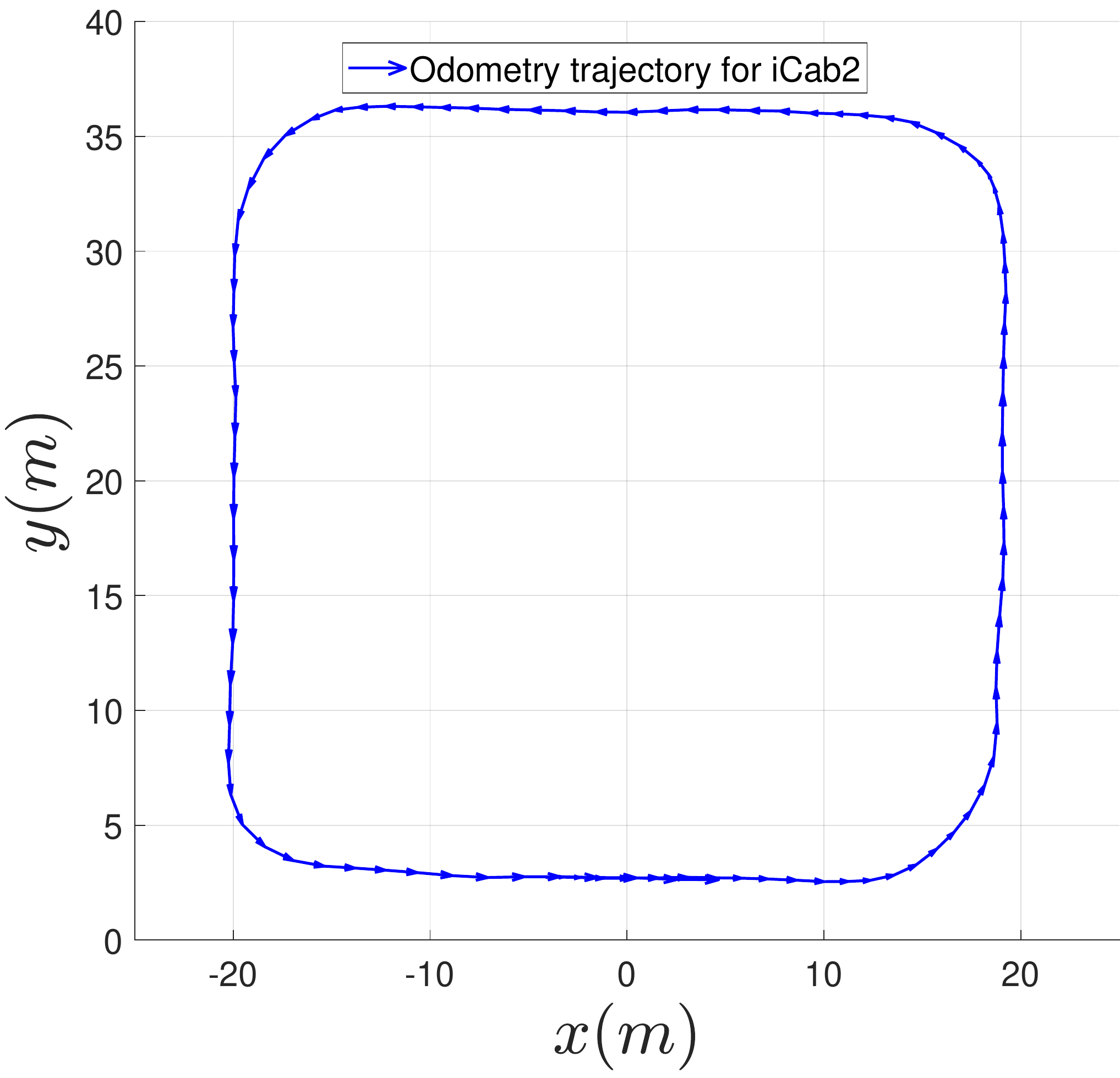}
		\caption{Perimeter monitoring}
		\label{fig:PM2}
	\end{subfigure}%
	\begin{subfigure}[t]{0.5\textwidth}
		\centering
		\includegraphics[height=5cm, width=6cm]{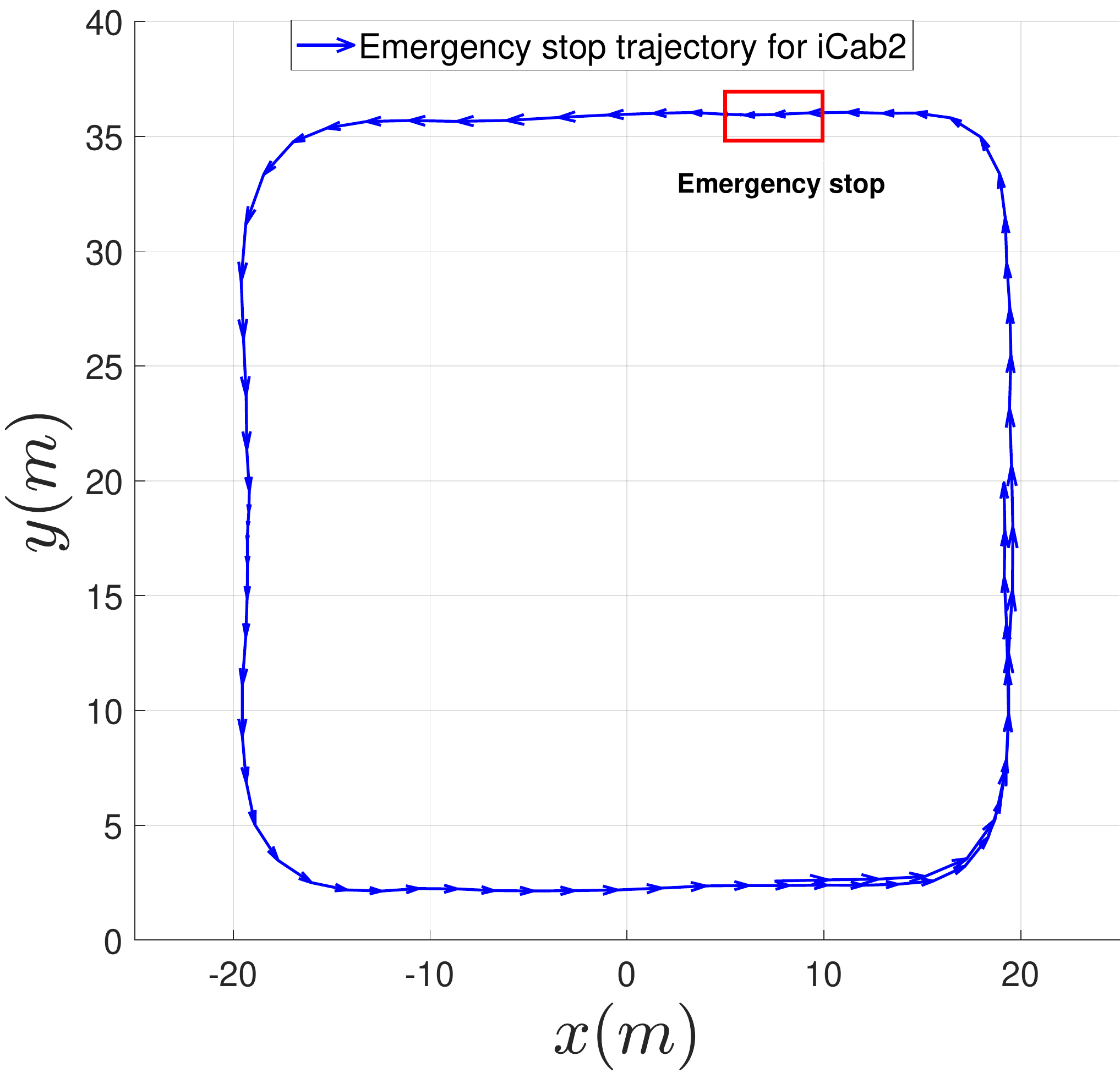}
		\caption{Emergency stop}
		\label{fig:ES2}
	\end{subfigure}
	\caption{Odometry data for iCab2}
	\label{fig:Position data Icab2}
\end{figure}

\begin{figure}
\centering
 	\includegraphics[width = 0.5 \linewidth ]{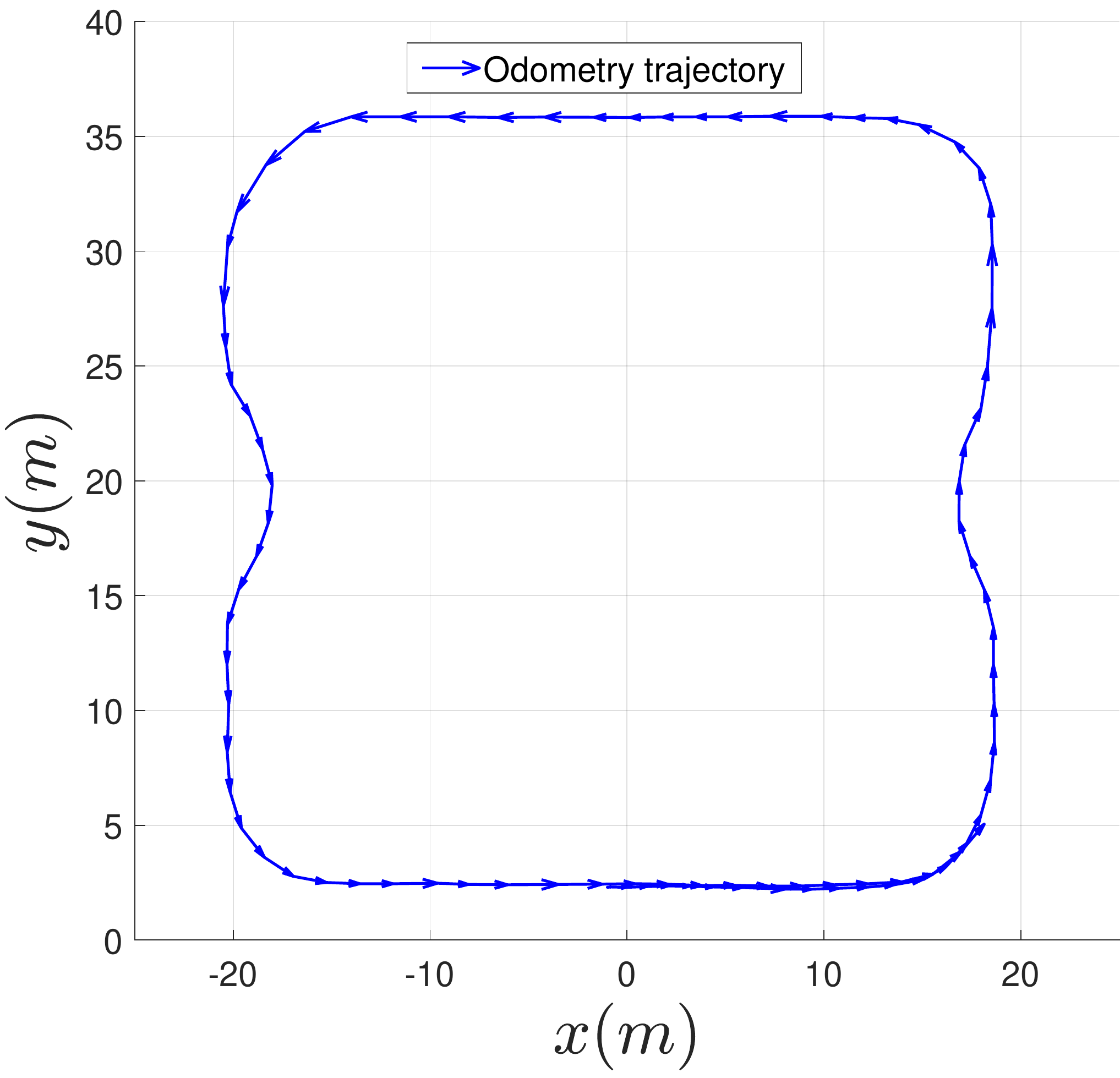}
\caption{Odometry data for pedestrian avoidance (iCab)}
\label{fig:Pedavoi}
\end{figure}

\begin{figure}
	\begin{subfigure}[t]{0.48\textwidth}
		\centering
		\includegraphics[height=5cm, width=6cm]{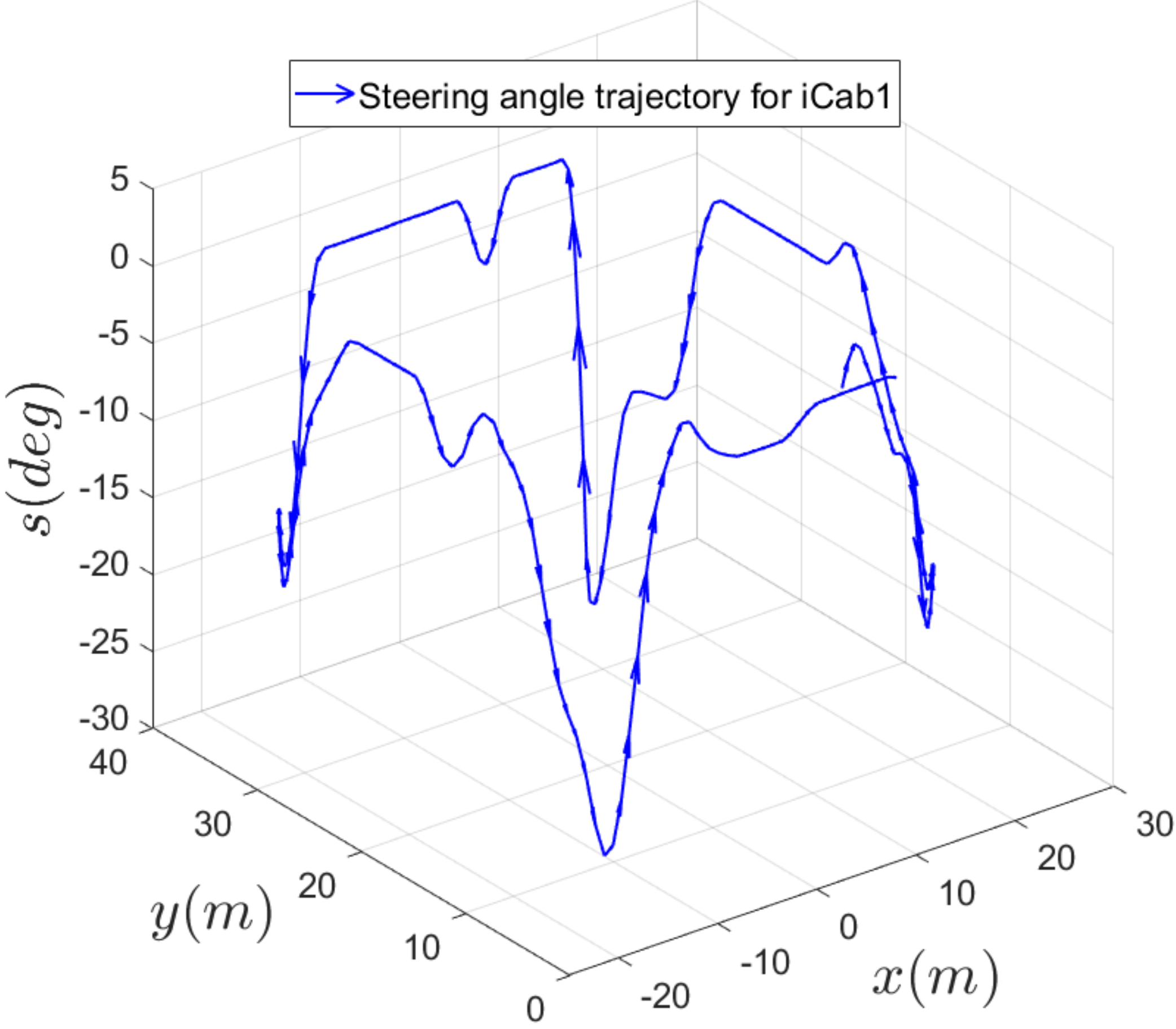}
		\caption{Steering angle(\textit{s}) w.r.t position}
		\label{fig:S}
	\end{subfigure}%
	\begin{subfigure}[t]{0.5\textwidth}
		\centering
		\includegraphics[height=5cm, width=6cm]{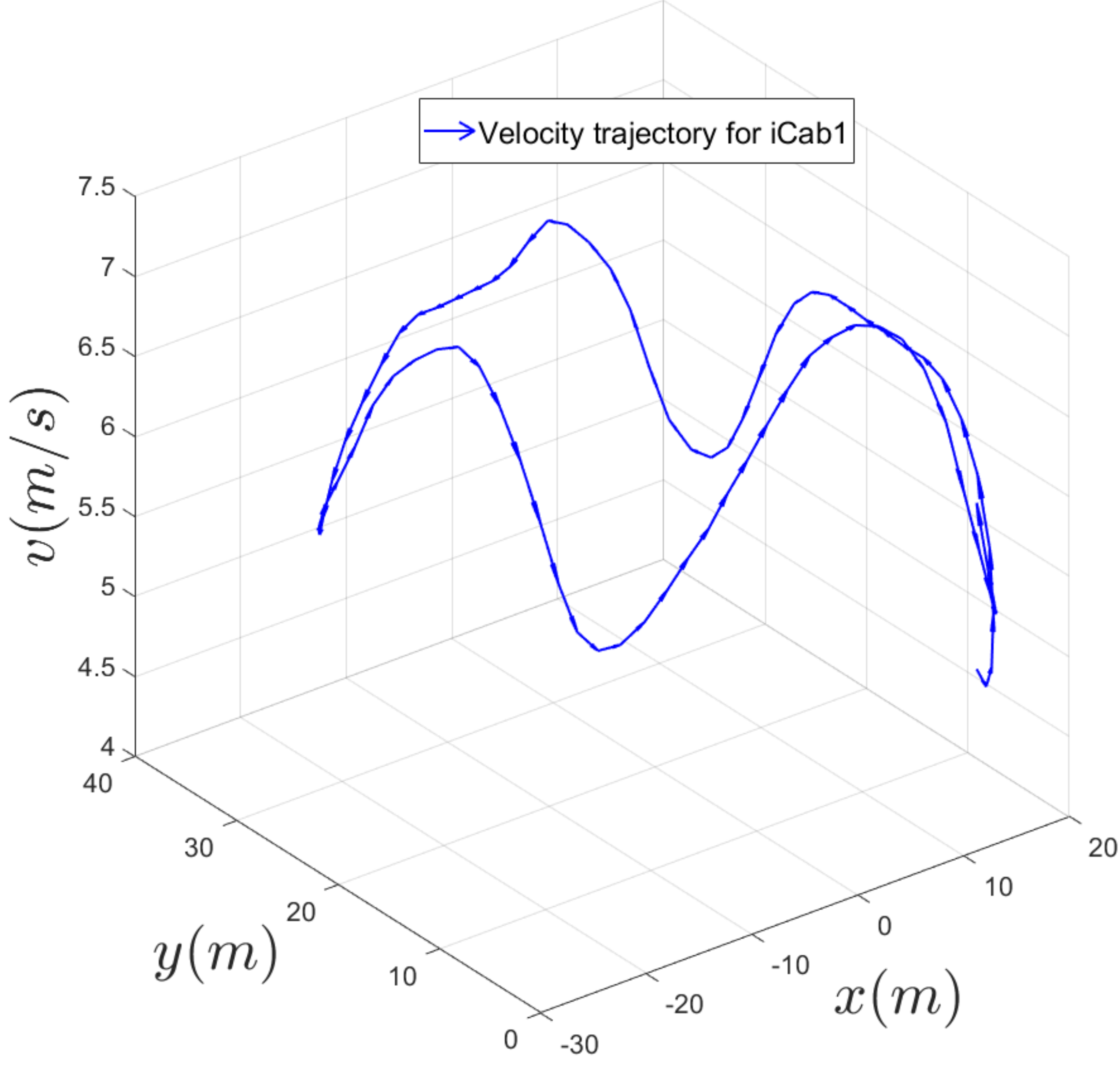}
		\caption{Velocity(\textit{v}) w.r.t position}
		\label{fig:V}
	\end{subfigure}
	\caption{Control data for iCab1 for perimeter monitoring task}
	\label{fig:Control data}
\end{figure}

\begin{figure}
\centering
 	\includegraphics[height=5cm, width=6cm ]{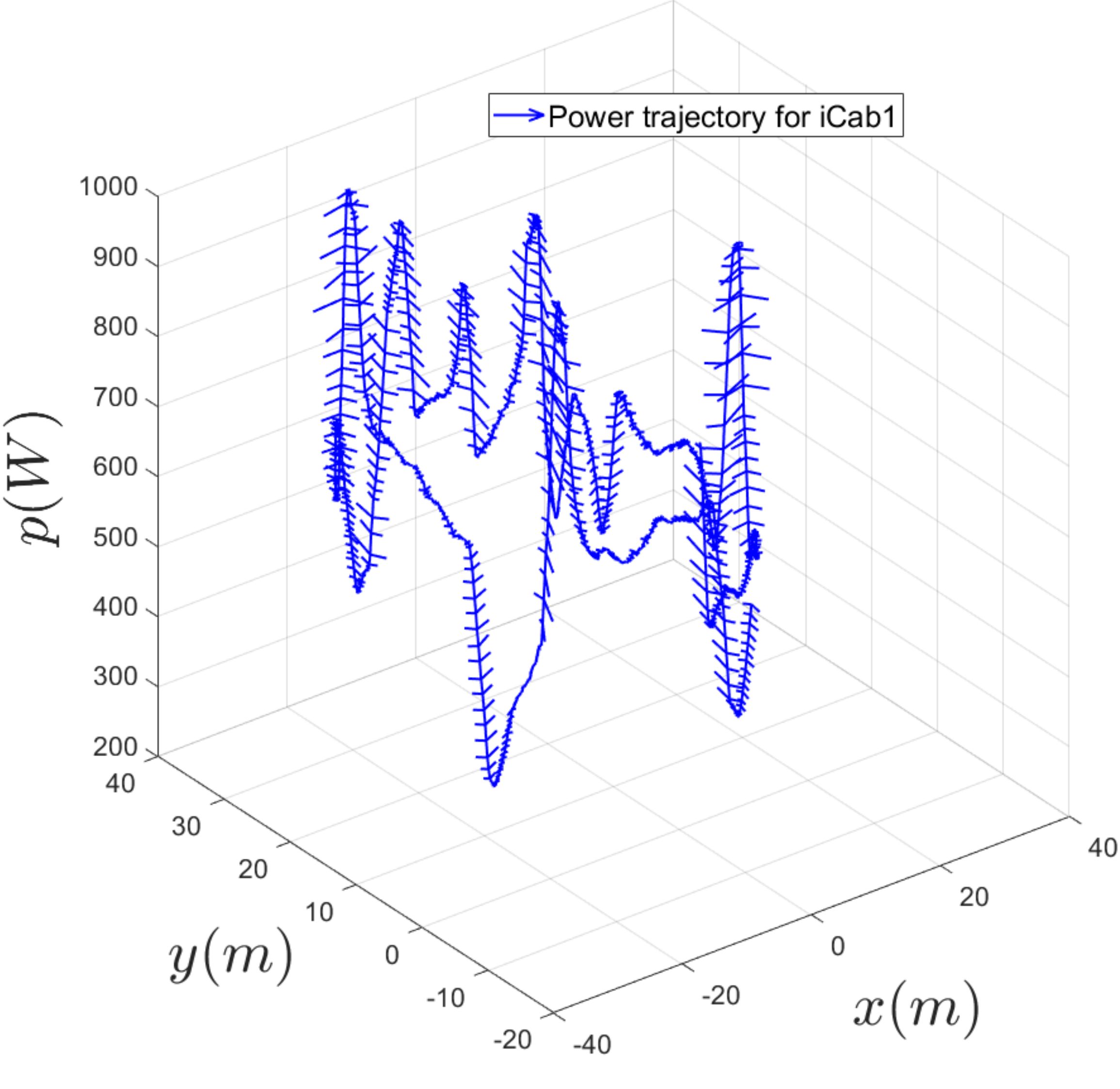}
\caption{Rotor power(\textit{p}) w.r.t position for iCab1 for perimeter monitoring task}
\label{fig:P}
\end{figure}

\subsection{Scenarios} \label{scenarios}
\begin{figure}
\centering
 	\includegraphics[width = 1\linewidth ]{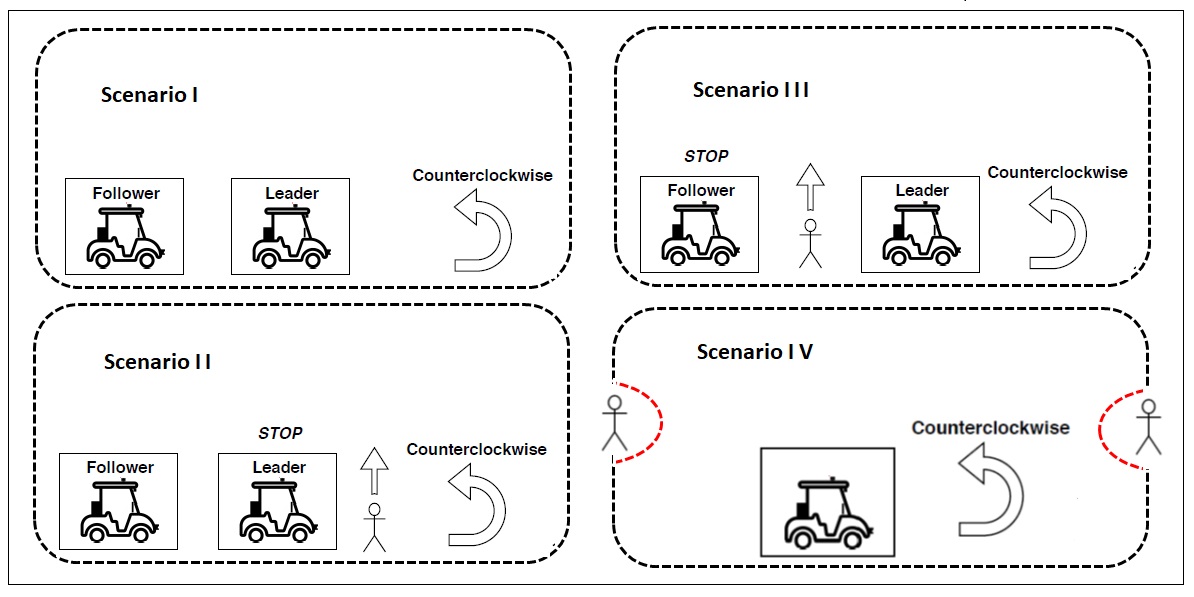}
\caption{Scenarios for PMT}
\label{fig:Scenarios}
\end{figure}

To generate the required data for the model learning and testing, we have conducted two types of experiments with four different scenarios shown in Fig. \ref{fig:Scenarios} and the description below.

\begin{itemize}
 \item \textbf{Scenario I}: Perimeter monitoring task (PMT)\\  The two iCab vehicles perform the platooning operation in a counterclockwise direction by following a rectangular trajectory in a closed environment (refer Scenario I in Fig. \ref{fig:Scenarios}), four laps in total. Fig.\ref{fig:PM1} and Fig.\ref{fig:PM2} show the plots of odometry data for the perimeter monitoring task for iCab1 and iCab2 respectively. Moreover, Fig.\ref{fig:S} and Fig.\ref{fig:V} show the steering angle ($S$) and rotor velocity($V$) data respectively of iCab1 plotted w.r.t it's own odometry position data. The rotor power data plotted w.r.t iCab1's position is illustrated in Fig.\ref{fig:P}. The paired data combinations from this scenario are used for model learning in the training phase.
    \item \textbf{Scenario II}: Emergency stop 1 (ES1)\\
    Both vehicles perform the same experiment of perimeter monitoring task (PMT), but now a random pedestrian crosses in front of the leader vehicle(i.e., iCab1) (refer to Scenario II in Fig. \ref{fig:Scenarios}). When the leader vehicle detects the dynamic obstacle (i.e., randomly crossing pedestrian), it automatically executes an emergency stop and waits until the pedestrian fully moves out from the danger zone. Meanwhile, the follower (i.e., iCab2) detects the leader's emergency brake action (by receiving the trajectory position data from iCab1 ) and mimics the same action of an emergency stop. Once the leader continues the PMT, the follower also starts moving and continues the  PMT. Fig.\ref{fig:ES1} and Fig.\ref{fig:ES2} show the plots of odometry data for the emergency stop criteria for iCab1 and iCab2 respectively. The intervals that the iCab vehicles perform an emergency stop is marked as a red box.
    \item \textbf{Scenario III}: Emergency stop 2 (ES2)\\
    In this scenario, a random pedestrian crosses in front of the follower(iCab2) vehicle, as shown in Fig \ref{fig:Scenarios} (Scenario III), while they perform the co-operative driving task of PMT. As soon as the follower vehicle detects the dynamic pedestrian's presence, it executes an emergency brake operation. However, the leader doesn't stop as in Scenario II; instead, it continues PMT. 
    \item \textbf{Scenario IV}: Pedestrian avoidance \\ 
    This scenario considers a standstill pedestrian appears along the path in two different locations that interfere with the perimeter monitoring task performs by an iCab vehicle. In this case, the pedestrian is a static obstacle. When a pedestrian appears in front of the vehicle, it executes an avoiding maneuver (the red part of the trajectory in Scenario IV in Fig.\ref{fig:Scenarios})  and continues its perimeter monitoring task.  The odometry data plot of the pedestrian avoidance scenario is shown in Fig.\ref{fig:Pedavoi}.
\end{itemize}

\section{Results}\label{section IV} 

The learned models have been tested with the data set collected from different co-operative tasks and a single-vehicle scenario task. The models' anomaly detection capability has compared to know the best pair-based feature of the vehicles. The overall training time (which includes synchronization, GNG clustering, vocabulary generation, and transition probability estimation) for all the four modalities was about $56$ seconds. On average, each modality consumed $14$ seconds for model learning, of the training data size of  $3200X2$. 
The computation time of the MJPF algorithm can mainly depend on the number of particles used and the test data size. The average calculation time of anomaly by the MJPF belongs to each pair based model for the test data size of $800X2$ (one complete lap) is 20 seconds in the test phase. Therefore, the average computation time for each abnormality sample is 0.025 seconds, includes the time for state prediction, anomaly estimation, and updation of states.

The abnormality threshold value is fixed based on the concept of validating the learned generative model. The prediction is distributed according to multivariate Gaussian, and the covariance matrix characterizes the shape of the function. If more percentage of the evidence(observation) falls under the prediction distribution, it is considered normal.  In this work, we have assumed that the generative model's prediction could be inside 60 \% of the distribution. If at least 50\% of the observation falls inside this confidence region, such a situation is considered normal. On the other hand, if more percentage of the observed evidence falls outside the 60\% of prediction, it is regarded as abnormal. By considering the above conditions, the anomaly threshold value is automatically identified and is 0.3.

The model testing phase consists of two parts: Phase I presented the results and analysis of the self-awareness functionality of the pair based DBN models and the initial level collective awareness functionality developed by the coupled DBN model. The coupled DBN model can be useful in situations like the pair based DBN model fails to detect the anomaly happens around other agents while performing co-operative tasks.

\begin{figure}
\centering
 	(a)\includegraphics[width=0.87\textwidth]{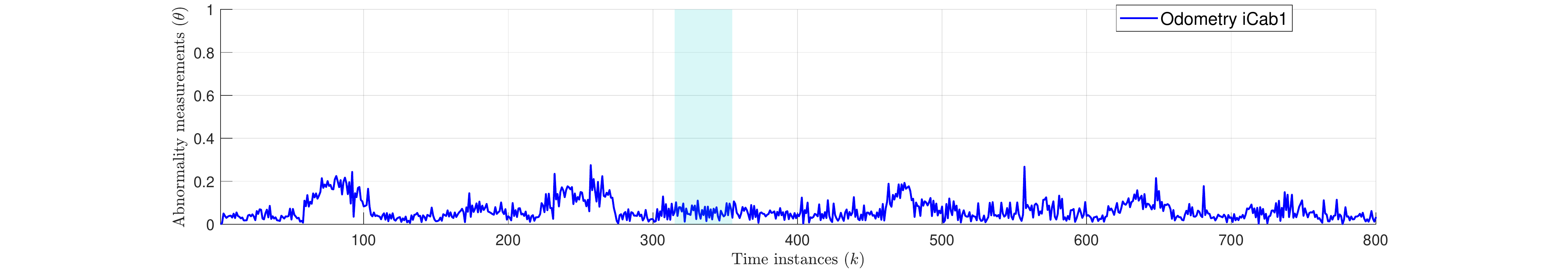}
 	(b)\includegraphics[width=0.87\textwidth]{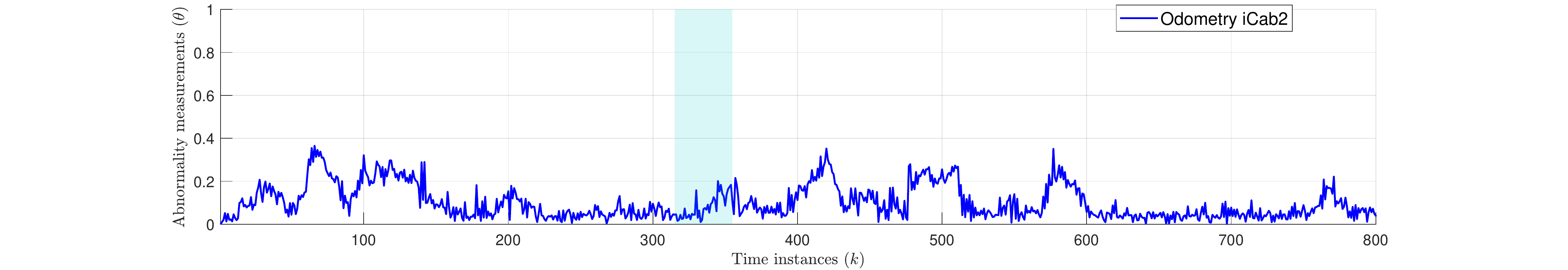}
\caption{Abnormality measurements for odometry: (a) $iCab1$, (b) $iCab2$}
\label{fig:XYcomb}
\centering
    (a)\includegraphics[width=0.87\textwidth]{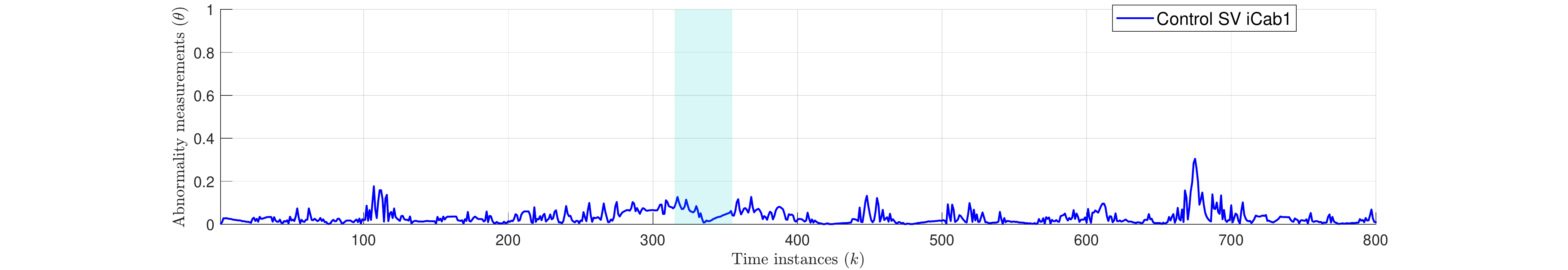}
 	(b)\includegraphics[width=0.87\textwidth]{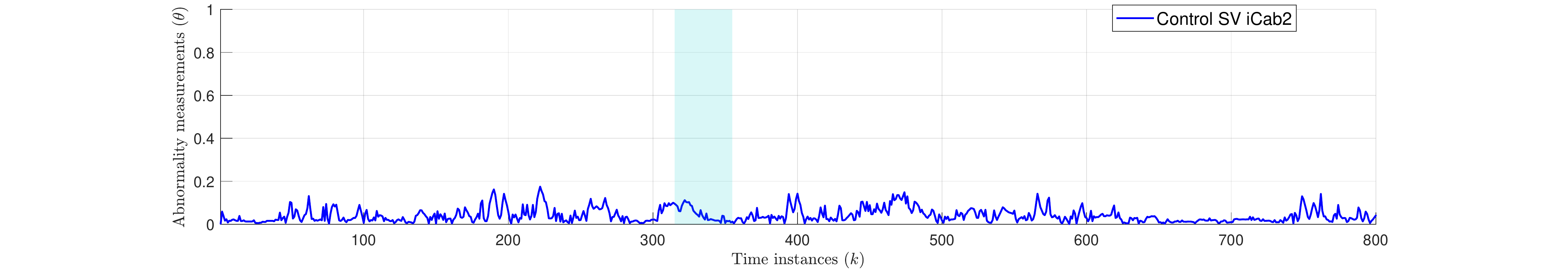}
\caption{Abnormality measurements for control (SV): (a) $iCab1$, (b) $iCab2$}
\label{fig:SVcomb}
\centering
    (a)\includegraphics[width=0.87\textwidth]{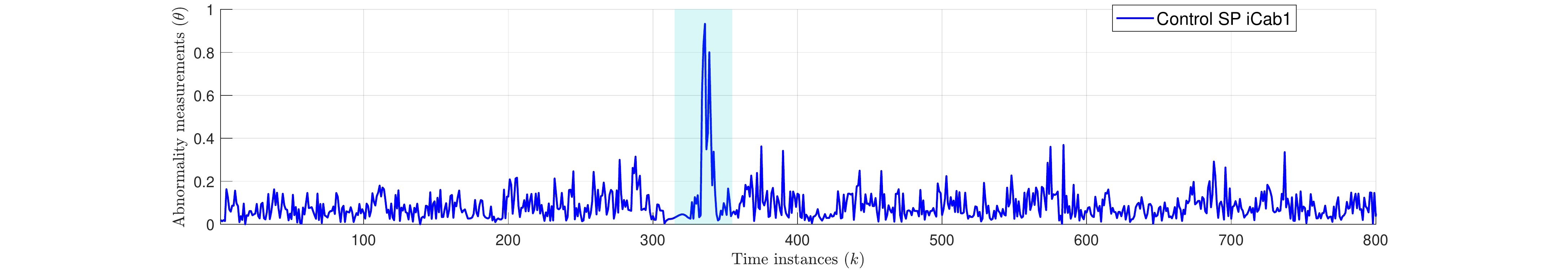}
 	(b)\includegraphics[width=0.87\textwidth]{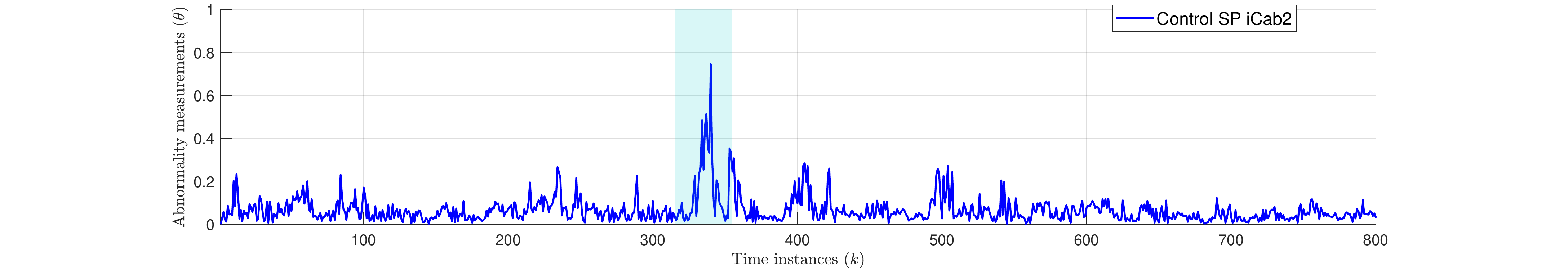}
\caption{Abnormality measurements for control (SP): (a) $iCab1$, (b) $iCab2$}
\label{fig:SPcomb}
\centering
    (a)\includegraphics[width=0.87\textwidth]{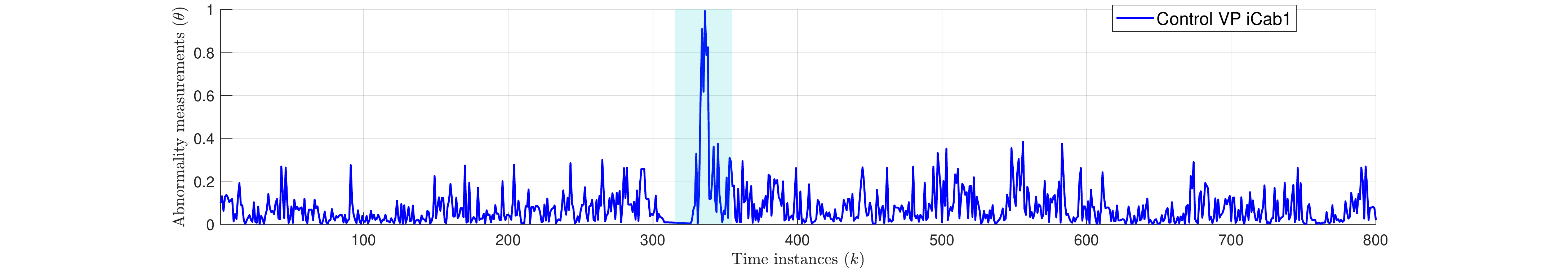}
 	(b)\includegraphics[width=0.87\textwidth]{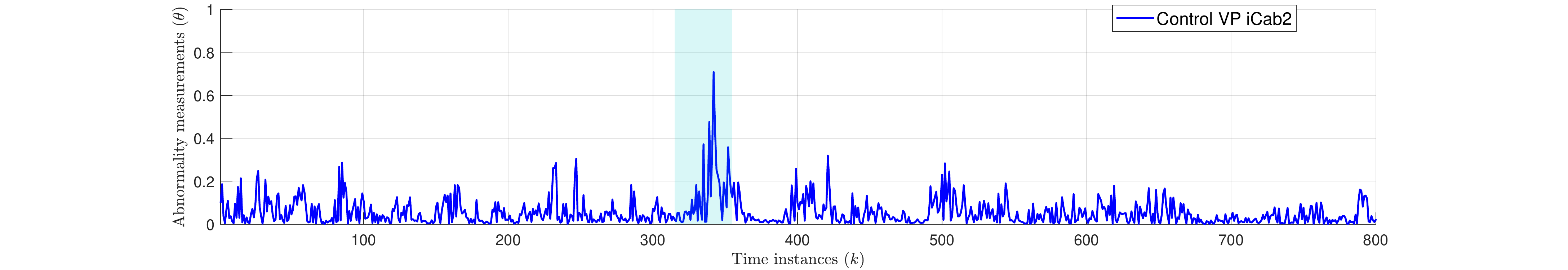}
\caption{Abnormality measurements for control (VP): (a) $iCab1$, (b) $iCab2$}
\label{fig:VPcomb}
\end{figure}

On the other hand, Phase II gave focus to test the models learned from single-vehicle experience, and the abnormality estimation results obtained from different pair based DBN models analyzed to check the best pair based feature. The \textit{cyan} shaded area from Fig.\ref{fig:XYcomb} to Fig.\ref{fig:pedavresults} shows the intervals where vehicles encountered static/dynamic obstacles.

\subsection{Test phase I }

In this phase, we used the data sets collected from two different co-operatives driving task scenarios. The results of Hellinger Distance(HD) abnormality measurements estimated by the DBN models learned from four different pair-based features of the vehicles presented. 
Moreover, the usefulness of coupled DBN models collective awareness functionality examined by estimating discrete level global anomaly.  

\subsubsection{Emergency stop scenario I } 
The anomaly estimation results by MJPF, whereas paired data sequences from Scenario II (described in Sec \ref{scenarios}), fed as input to MJPF presented in this part.  The results from each pair-based DBN models and the coupled DBN model learned from odometry position data are described below.

\begin{enumerate}[label=(\roman*)]
  \item \textbf{Odometry ($X-Y$)}: The switching DBN model in this work is designed for the control part of the vehicles. However, we have considered odometry data and tested the performance of the learned DBN. Fig. \ref{fig:XYcomb} shows the plots of abnormality measures by examining odometry data for the vehicle leader (iCab1) and the vehicle follower (iCab2), respectively. During the interval (cyan shaded area) while a random pedestrian crosses in front of the leader vehicle (icab1), and it stops, there isn’t any significant difference in HD value for iCab1 (leader) as well as iCab2 (follower). This behavior is because, during that interval, the vehicles are always inside the normal trajectory range. \\
 However, specific intervals when the vehicles deviate from the normal range with respect to the trajectory data used to learn models; the HD measures provided a value of about 0.2 during those intervals. It means that the odometry DBN model was able to predict if any trajectory deviation occurred.
\item \textbf{Steering angle-Velocity ($S-V$)}: It is necessary to check different pairs of 2D sensory data combinations to know the vehicles' best pair-based features. The models learned from the steering-velocity pair doesn't notice the abnormal situation happens in the environment.  This is an expected result as the vehicles' steering values don't change much when they perform an emergency stop.
The HD abnormality plots for iCab1 and iCab2 vehicles are depicted in Fig.~\ref{fig:SVcomb}, not detecting pedestrians' presence. So that, this pair is not considered a useful feature for the abnormality detection purpose for similar scenarios.
 \item \textbf{Steering angle-Power ($S-P$)}: When a pedestrian crosses in front of the leader vehicle (iCab1), the HD value is shown high during that interval, as shown in  Fig.~\ref{fig:SPcomb} (a). The follower(iCab2) receives the shared trajectory data from the leader, understands the presence of anomaly, and performs an emergency stop operation to keep a minimum distance with the leader. So that the DBN model inside the follower also detects the anomaly, and the HD value becomes high in that interval, as shown in Fig.~\ref{fig:SPcomb} (b).  Thus $S-P$ is a useful feature of the vehicle in detecting an anomaly for similar tasks.

  \item \textbf{Velocity-Power ($V-P$)}: The last pair tested is velocity and power consumption, which are highly related. In Fig.~\ref{fig:VPcomb} (a), the HD anomaly when a pedestrian cross in front of the leader is shown in cyan color.  The highest HD values(closer to $1$) are shown in this modality w.r.t other modalities. For the follower vehicle, the abnormality measurement is very significant, as shown in Fig.~\ref{fig:VPcomb} (b) due to the brake operation performed by following the action of the leader vehicle. The consecutive peaks(that are closer to 0.3) in the plots are caused by the high acceleration when the leader vehicle starts moving, but the current distance between them is still lower than the desired.

 \end{enumerate}
 To summarize, the pair-based DBN learned from $S-P$ and $ V-P $ data were able to predict the unusual situations present; however, odometry and $ S-V $ pairs of control did not show the right combination to detect the agents abnormal behavior. 
 
 \subsubsection{Emergency stop scenario II } 
This part analyzes the anomaly signal estimation by the models considering the sensory data of Scenario III (refer Sec \ref{scenarios}). In this special case scenario, when a pedestrian crosses in front of the follower(iCab2) vehicle, it performs the emergency stop, and the leader vehicle continues the PMT. The DBN model inside the leader doesn't show any anomaly; only the DBN models inside the follower vehicle detect pedestrians' presence.  
  
In such situations, the coupled DBN model plays an important role. The model learned from both vehicle's position data represents the collective situation and estimates anomaly that considers both vehicle's behavior.  The model will perform a discrete level anomaly estimation based on the co-occurrence probability matrix that infers the collective awareness information. For simplicity, we have only presented the results from $S-P$ modality to show the self-awareness functionality and odometry modality for collective self-awareness.  
 
 \begin{figure}
\centering
 	(a)\includegraphics[width=0.87\textwidth]{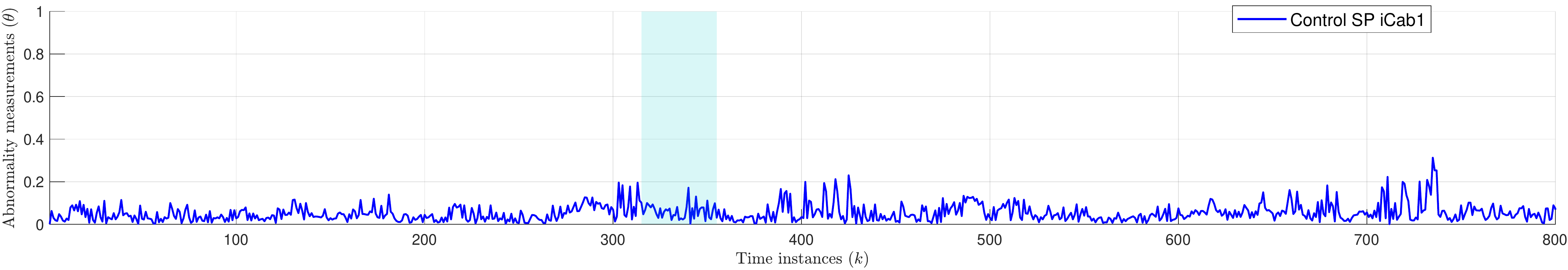}
 	(b)\includegraphics[width=0.87\textwidth]{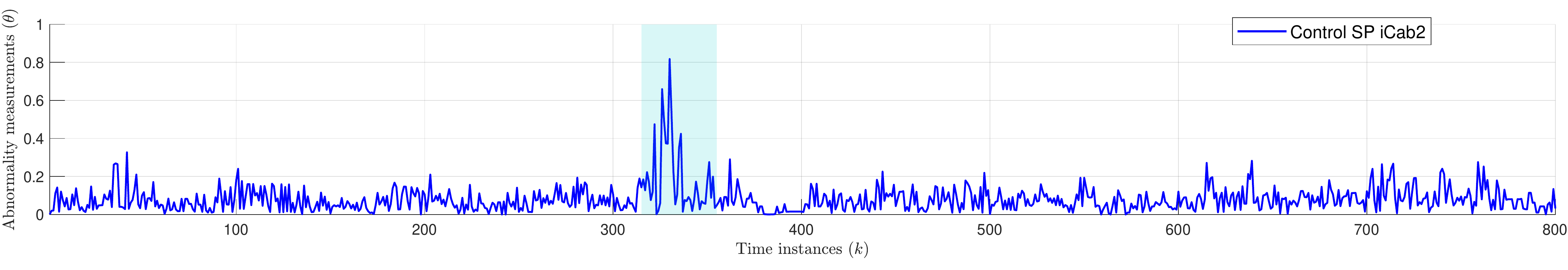}
\caption{Abnormality measurements for control SP: (a) $iCab1$, (b) $iCab2$}
\label{fig:SPabnsceII}
\end{figure}

 \begin{enumerate}[label=(\roman*)]
  \item \textbf{Pair based Steering angle-Power ($S-P$) model}: The HD anomaly measurements for iCab1 and iCab2 for Steering-power ($S-P$) modality is shown in Fig.\ref{fig:SPabnsceII} (a) and Fig.\ref{fig:SPabnsceII} (b) respectively. Contrary to the previous case (Emergency stop scenario I), the HD anomaly indicator of the DBN model inside the leader(iCab1) vehicle doesn't show any high peaks as the vehicle doesn't stop anywhere during the PMT, and this is an expected result. Simultaneously, the DBN models inside the follower((iCab2) vehicle detect abnormality when the vehicle executed emergency brake operation during the dynamic pedestrian crosses in front of it. We haven't included the results of models learned from other pairs and assumed that results would be are similar to the previous shown co-operative scenario.

  \item \textbf{Collective self-awareness: Odometry ($X-Y$)}: The situations in which the pair based self-awareness models fail to detect anomaly around other agents, the collective awareness model (i.e., coupled DBN) plays a vital role. For instance, the pair based DBN model of $S-P$ modality inside iCab1 vehicle , tested with the data sets of Scenario III (refer Sec \ref{scenarios}), didn't show any high peaks in estimated Hellinger Distance(HD) values. Therefore the leader vehicle (iCab1) was not aware of the anomaly happening around the follower vehicle (iCab2).
  \begin{figure}
\centering
 	\includegraphics[width = 1\linewidth ]{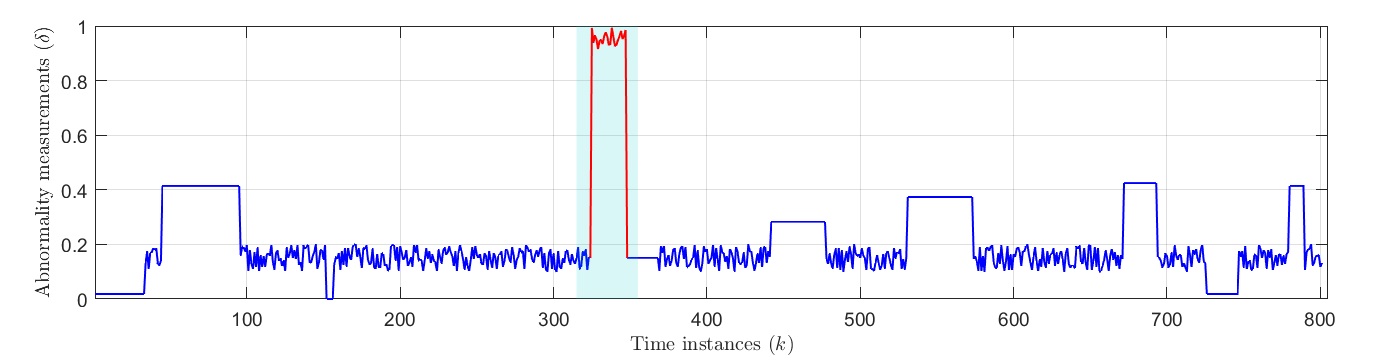}
\caption{Abnormality measurements based on coupled co-occurrence matrix}
\label{fig:cooccSceII}
\end{figure}
 In this situation, the collective awareness model inside iCab1 plays an essential role in showing anomaly present around iCab2. The coupled DBN model inside iCab 1 was able to estimate the anomaly (presence of dynamic pedestrian) happens around the follower vehicle.  The discrete probabilistic anomaly signal shows high values, as in Fig.\ref{fig:cooccSceII} during the interval when the pedestrian is inside the danger zone. The collective DBN model inside any of the vehicles can track the anomaly happens around any of the other vehicles that are part of the co-operative driving task. 
\end{enumerate}
 
 \subsection{Test phase II :Pedestrian avoidance scenario}

In this part, the abnormality measurements obtained when the model tested with the data from a pedestrian avoidance task (refer to Scenario IV in Fig.\ref{fig:Scenarios} ) have been presented. A total of four pairs of data combinations were examined and described below. 

\begin{figure}
\centering
 	(a)\includegraphics[width=\textwidth]{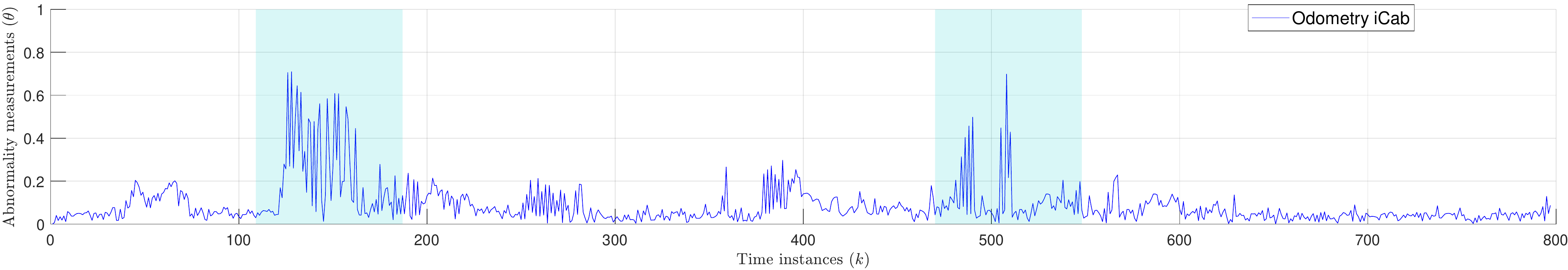}
 	(b)\includegraphics[width=\textwidth]{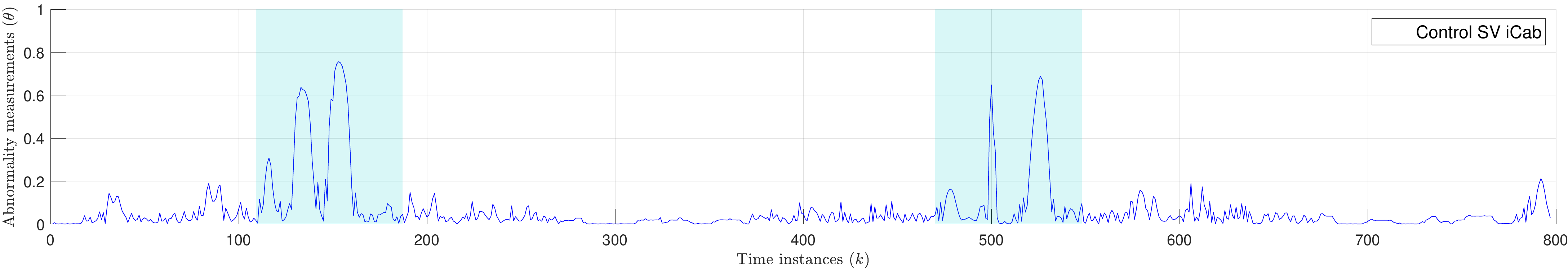}
 	(c)\includegraphics[width=\textwidth]{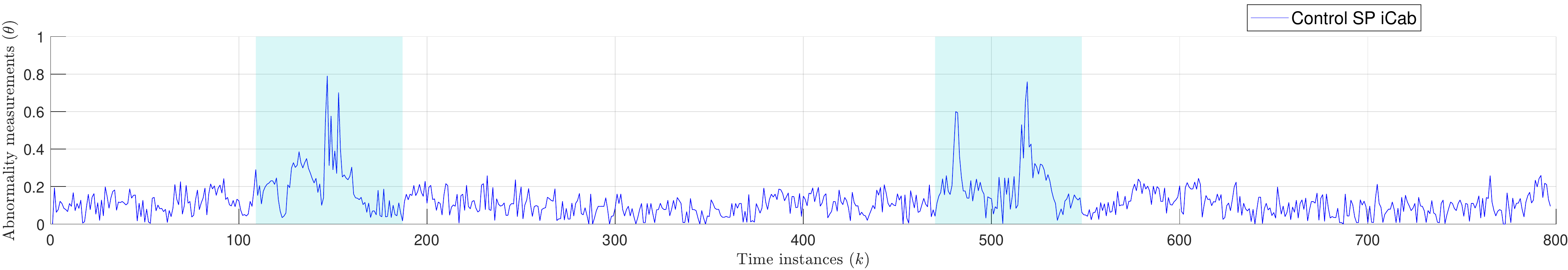}
 	(d)\includegraphics[width=\textwidth]{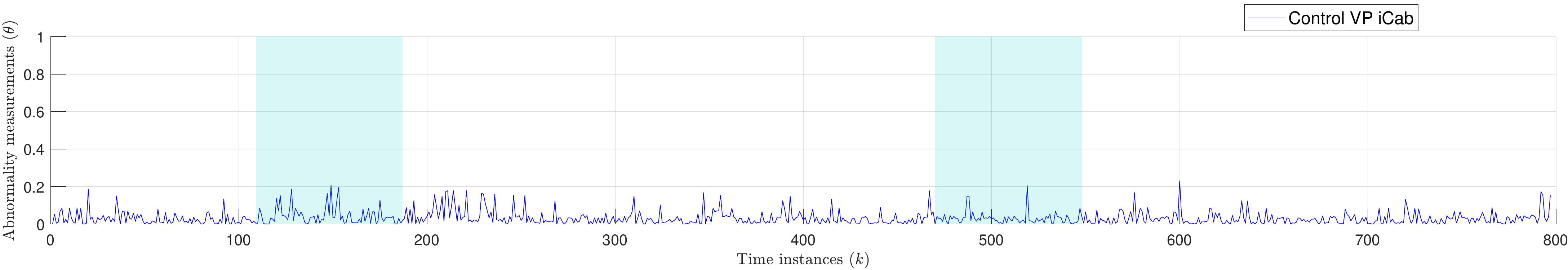}
\caption{Abnormality measurements for pedestrian avoidance: (a) $Odometry\:XY$, (b) $Control\:SV$ (c) $Control\: SP$ (d) $Control\: VP$}
\label{fig:pedavresults}
\end{figure}

\begin{enumerate}[label=(\roman*)]
\item \textbf{Odometry ($X-Y$)}: 
    The HD abnormality measurements from the odometry test data of scenario IV plotted in Fig.~\ref{fig:pedavresults}(a). The model inside the iCab vehicle detects abnormality during the vehicle performs avoidance maneuver by detecting the presence of a static pedestrian in front of it. As the trajectory deviates from the normal one, the abnormality peaks are high as expected.
   \item \textbf{Steering Angle-Velocity ($S-V$)}: When used $ SV $ pair test data from scenario IV, the HD abnormality measurements are shown in Fig.~\ref{fig:pedavresults}(b). Like the odometry data, the model learned from the $ S-V $ pair also shows abnormality peaks during those intervals. This high value in the HD metric is because when a pedestrian appears, the vehicle takes the trajectory (shown as the red dotted line in Fig.\ref{fig:Scenarios} ) that is different than the one in the training phase, so that there would be significant deviations in the steering angle of the vehicle.
 
   \item \textbf{Steering Angle-Power ($S-P$)}: 
    Fig.~\ref{fig:pedavresults}(c) shows the HD measurement plot of $SP$ pair. When the vehicle encounters a pedestrian in two different locations, it shows high peaks in the HD plot (cyan shaded area) during those intervals. Other small spikes indicated the sensor noise and abnormalities in the environment. 
   
   \item \textbf{Velocity-Power ($V-P$)}: 
    The HD abnormality measurements from the $VP$ pair test data for the considered vehicles plotted in Fig.~\ref{fig:pedavresults}(d). Here, the HD measures do not show any peaks during those intervals because the considered vehicle's velocity and power don't change much when they execute the avoidance maneuver. 

\end{enumerate}

The pair based models inside the vehicles play an essential role in detecting anomaly(presence of dynamic and static obstacles) happens around itself, and also some instances of co-operative scenarios. On the other hand, the collective awareness model (coupled DBN) detects anomaly occurs around any of the vehicles that are part of co-operative driving tasks by exploiting the data shared by the communication scheme.

\section{Conclusion and future work } 
\label{section V1}
This paper proposed a method to develop self-awareness models considering pair based features of the vehicles. To learn the switching DBN models, low dimensional multisensory data describe the normal behavior of the vehicles used. Such a learned data-driven DBN model can automatically detect abnormal situations instead of defining the metric's upper and lower limits. A set of DBN models learned inside each entity by considering pair based features helps confirm the best model that fits to detect abnormalities. The obtained results from each of the pair based DBN models show that our method works well in detecting the dynamic anomalies in the surrounding environment of the vehicle. We have considered data sets from different co-operative driving scenarios as well as a single-vehicle scenario. A performance comparison was performed of the models learned from various pair-based features of the vehicles. \\
Additionally, an initial level collective awareness model(i.e., coupled DBN ) is proposed to detect collective anomaly when agents perform co-operative tasks. When the anomaly happens around one vehicle, the coupled DBN models inside other vehicles can detect the presence of anomaly. However this work needs additional work in future to improve the collective awareness functionality. 
The future work can include efficient communication schemes between the objects to share essential data among the vehicles to exploit collective awareness by considering different modalities. Such models can make the mutual prediction of the future states of the objects involved in the task. The model performance under the influence of wireless communication channels and the different communication protocols standards, environmental conditions, etc. need to be carefully studied. Additionally, the classification of detected abnormality by considering different test scenarios and comparing abnormality detection performance using other abnormality metrics could be considered.

\section*{Acknowledgement}
Research supported by the Spanish Government through the CICYT projects (TRA2016-78886-C3-1-R and RTI2018-096036-B-C21), Universidad Carlos III of Madrid through  (PEAVAUTO-CM-UC3M)  and the Comunidad de Madrid through SEGVAUTO-4.0-CM (P2018/EMT-4362). We gratefully acknowledge the support of NVIDIA Corporation with the donation of the GPUs used for this research.
\\


\begingroup
\let\clearpage\relax
\bibliographystyle{styles/bibtex/spbasic_unsrt}
\bibliography{references}
\endgroup

\end{document}